\documentclass[10pt,twocolumn,letterpaper]{article}

\usepackage[pagenumbers]{cvpr} 

\usepackage{tikz}
\usetikzlibrary{shapes.geometric, arrows.meta, positioning, calc}
\usepackage{tikz}
\usepackage{booktabs}
\usetikzlibrary{
    positioning,    
    shapes.geometric, 
    arrows.meta,    
    calc           
}
\usepackage{amsmath}  
\usepackage{amssymb}  

\usepackage[export]{adjustbox}

\usepackage{subcaption}

\usepackage{times}
\usepackage{latexsym}
\usepackage{bm}

\usepackage{url}
\usepackage{colortbl}
\usepackage[table]{xcolor}  

\usepackage{multirow}

\definecolor{cvprblue}{rgb}{0.21,0.49,0.74}
\usepackage[pagebackref,breaklinks,colorlinks,allcolors=cvprblue]{hyperref}

\def\paperID{20203} 
\def\confName{CVPR}
\def\confYear{2026}

\title{CaReFlow: Cyclic Adaptive Rectified Flow for Multimodal Fusion}

\author{
\Large Sijie Mai\thanks{Corresponding Author} \ \ \ \ \ Shiqin Han\\[0.2cm]
\Large South China Normal University\\
{\tt\normalsize \{sijiemai, 20222121019\}@m.scnu.edu.cn}
}

\begin{document}
\maketitle
\begin{abstract}
Modality gap significantly restricts the effectiveness of multimodal fusion. Previous methods often use techniques such as diffusion models and adversarial learning to reduce the modality gap, but they typically focus on one-to-one alignment without exposing the data points of the source modality to the global distribution information of the target modality. To this end, leveraging the characteristic of rectified flow that can map one distribution to another via a straight trajectory, we extend rectified flow for modality distribution mapping. Specifically, we leverage the `one-to-many mapping' strategy in rectified flow that allows each data point of the source modality to observe the overall target distribution. This also alleviates the issue of insufficient paired data within each sample, enabling a more robust distribution transformation. Moreover, to achieve more accurate distribution mapping and address the ambiguous flow directions in one-to-many mapping, we design `adaptive relaxed alignment', enforcing stricter alignment for modality pairs belonging to the same sample, while applying relaxed mapping for pairs not belonging to the same sample or category. Additionally, to prevent information loss during distribution mapping, we introduce `cyclic rectified flow' to ensure the transferred features can be translated back to the original features, allowing multimodal representations to learn sufficient modality-specific information. After distribution alignment, our approach achieves very competitive results on multiple tasks of multimodal affective computing even with a simple fusion method, and visualizations verify that it can effectively reduce the modality gap.
\end{abstract}    
\section{Introduction}
\label{sec:intro}


\begin{figure}
    \centering
    \begin{subfigure}[b]{0.49\linewidth} 
        \centering
        \includegraphics[width=\linewidth]{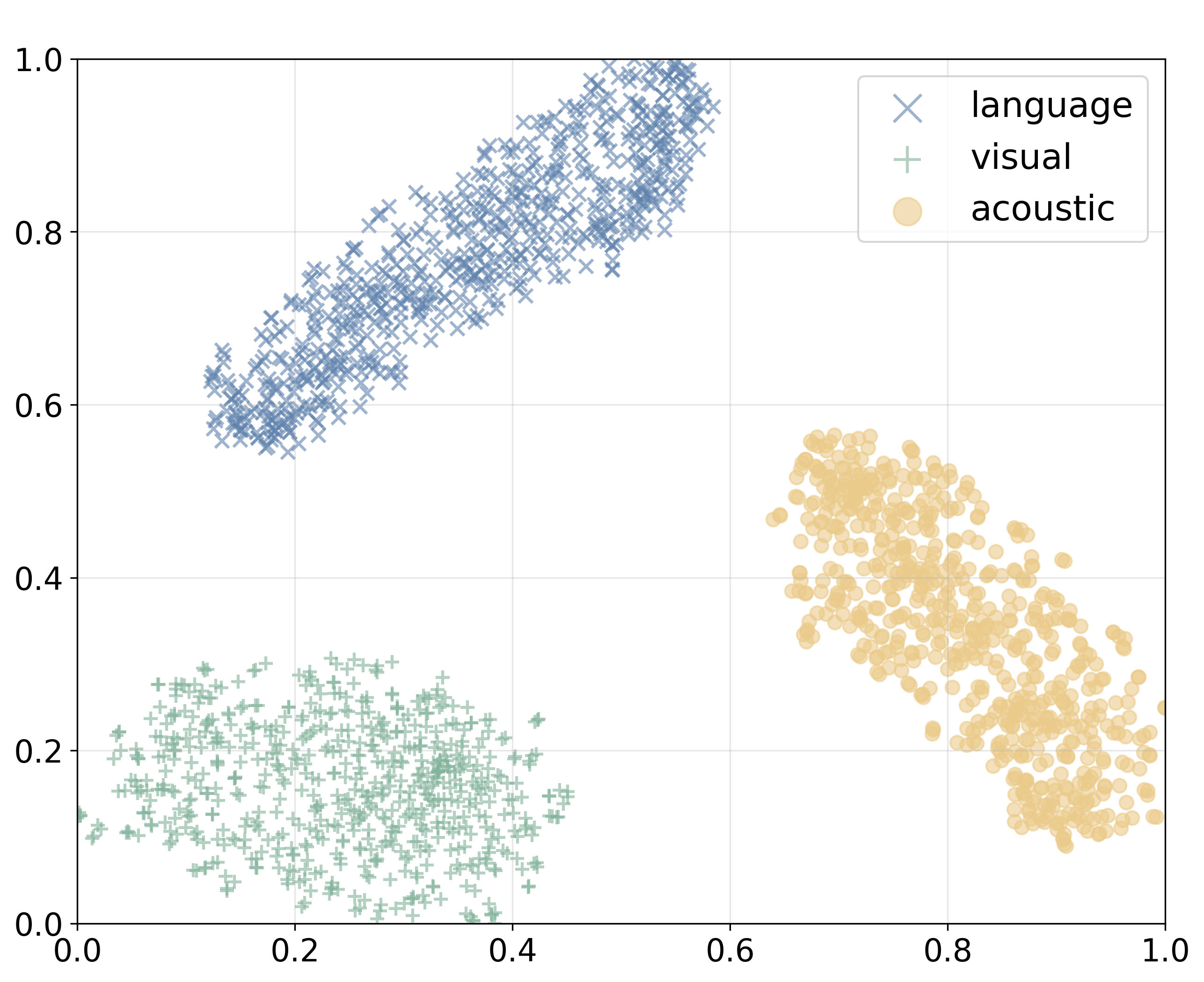}
        \caption{Modality gap}
    \end{subfigure}
    \begin{subfigure}[b]{0.49\linewidth}
        \centering
        \includegraphics[width=\linewidth]{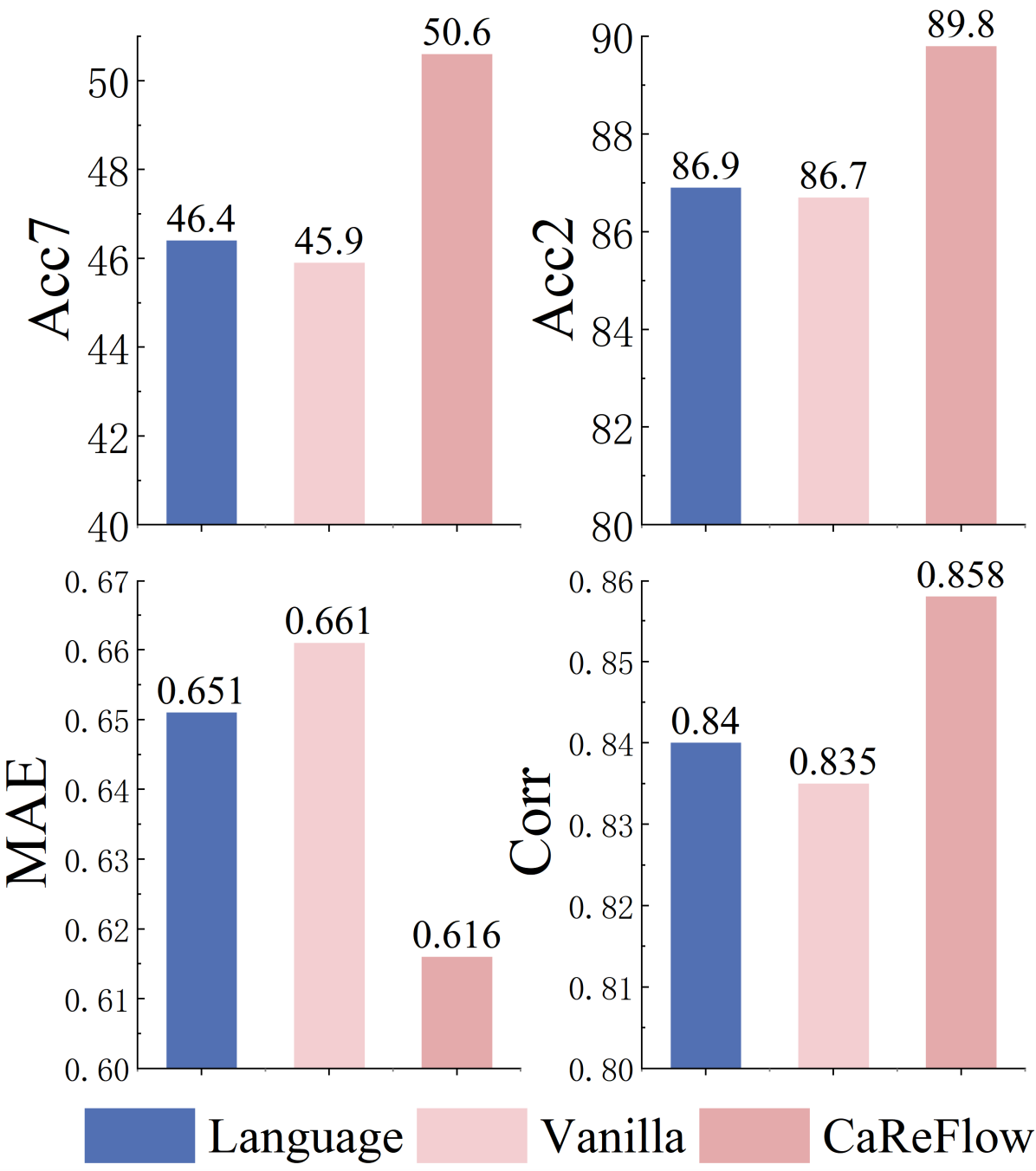}
        \caption{Results on CMU-MOSI \cite{zadeh2016multimodal} }
    \end{subfigure}
    \vspace{-0.2cm}
    \caption{(a) shows that there are significant differences in the distribution of different modalities in the feature space, and (b) show that vanilla multimodal model (using feature concatenation and MLP networks for fusion) even underperforms language-only model. After applying our CaReFlow to reduce modality gap, multimodal model significantly outperforms language-only model.}
    \label{vis_intro}
    \vspace{-0.3cm}
\end{figure}

Multimodal affective computing (MAC) has emerged as a pivotal area within the field of human-centered artificial intelligence, focusing on the analysis and interpretation of human sentiments and affective states through visual, acoustic, and language modalities \cite{li2023decoupled_cvpr,AffectiveComputing,8269806}. 
A core challenge in MAC lies in fusing complementary information from heterogeneous modalities. However, a significant obstacle, known as the `modality gap', severely impedes this fusion process \cite{ARGF,lee2025diffusion}. As shown in Fig.~\ref{vis_intro} (a), modality gap refers to the fundamental distributional discrepancy where data from different modalities (\eg, facial expressions in video and corresponding speech prosody) reside in distinct, non-aligned regions of the feature space due to the heterogeneous nature of different modalities and distinct feature extractors. Consequently, vanilla multimodal models struggle to model their complex inter-dependencies, leading to suboptimal results and poor generalization (see Figure~\ref{vis_intro} (b)).

Existing efforts to bridge modality gap have largely relied on techniques such as contrastive learning \cite{hycon,yang2024clgsi}, multimodal transformer \cite{MULT}, adversarial generative networks (GANs) \cite{ARGF}, and diffusion models \cite{imder,lee2025diffusion}. Despite their potential, these approaches often focus on pairwise one-to-one alignment between modalities. They typically do not expose a data point of source modality (\eg, a visual feature) to the `global distributional context' of the target modality during modality alignment. This narrow view restricts the model's ability to learn a holistic and robust modality alignment, especially when paired data is scarce. 

\begin{figure}[t]
\centering
\includegraphics[width=0.95\linewidth]{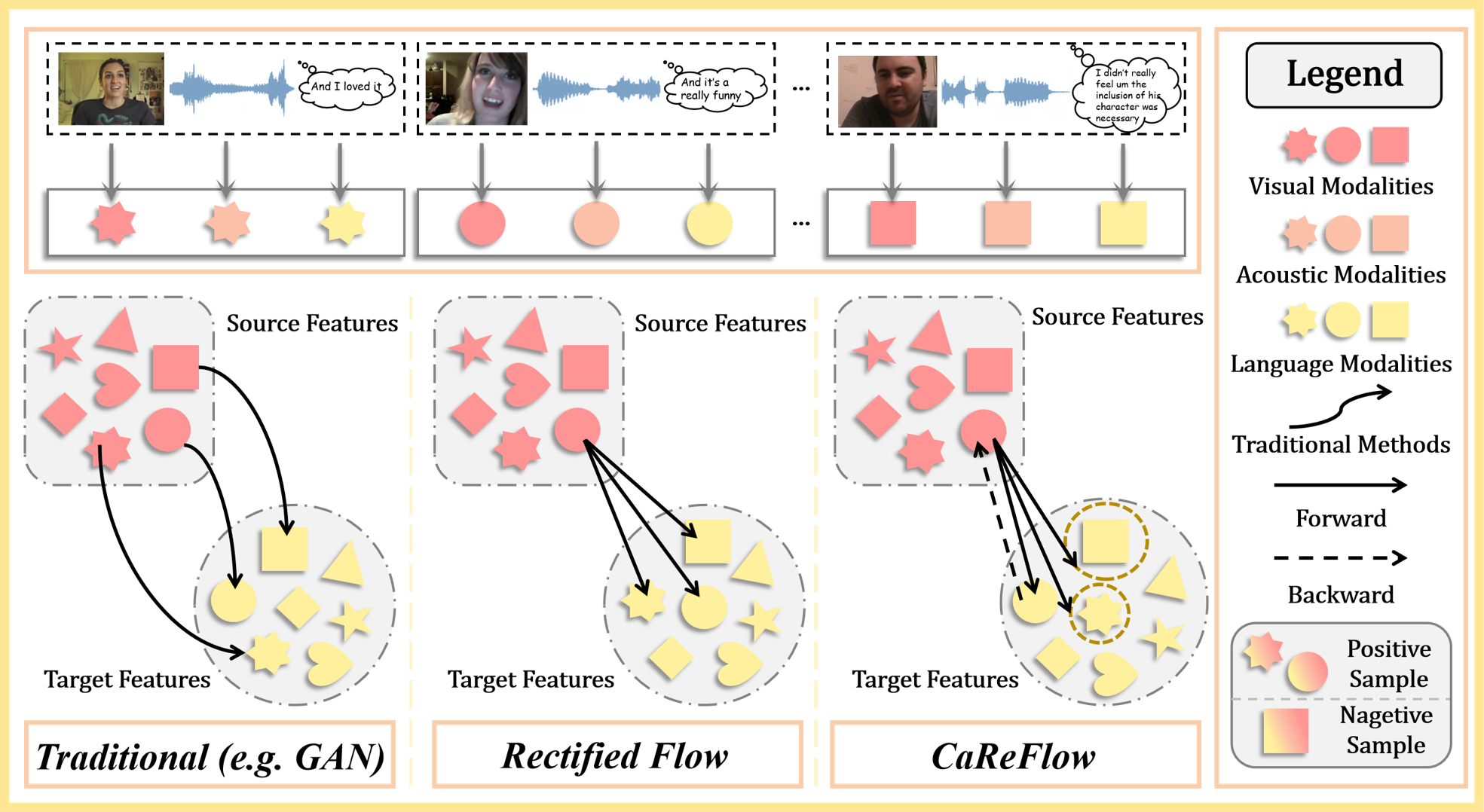}
\vspace{-0.2cm}
\caption{The comparison between tradition methods, rectified flow \cite{liuflow} , and our CaReFlow. CaReFlow implements one-to-many mapping, adaptive relaxed alignment, and cyclic information flow to improve the effect of modality alignment.}
\label{fig:intro}
\vspace{-0.3cm}
\end{figure}


Inspired by the recent success of rectified flow \cite{liuflow,li2025omniflow,lee2024improving}, a simple and effective generative framework that can learn a straight and fast trajectory to map one distribution to another, we adapt it for modality distribution alignment and propose the cyclic adaptive rectified flow (CaReFlow) framework. CaReFlow 
trains a drift force (velocity vector-field) model to straightly transform source distribution to target distribution, thereby explicitly and geometrically bridging the modality gap.
CaReFlow leverages `\textbf{one-to-many mapping}' strategy in rectified flow during the transformation (i.e., the rectification process) from source to target modality. As shown in Figure~\ref{fig:intro}, instead of pushing a source data point towards a single target, CaReFlow allows it to be influenced by the broader target modality distribution. This mapping strategy enables each source data point to observe the overall target distribution, facilitating a more effective alignment and mitigating the issue of insufficient paired data between modalities within each sample.

Furthermore, as presented in Figure~\ref{fig:intro}, CaReFlow designs two major innovations over the vanilla rectified flow paradigm to suit the unique challenges of multimodal distribution mapping. Firstly, considering the existence of inherent one-to-one modality correspondence within each multimodal sample, CaReFlow implements an `\textbf{adaptive relaxed alignment}' mechanism. It enforces stricter rectification for pairs belonging to the same sample, while applying a relaxed constraint for modality pairs from different samples or categories. The relaxed degree is adaptively determined by the label difference between two samples (pairs with same/similar labels should be aligned more closely in the feature space). Adaptive relaxed alignment helps to achieve a more precise and semantically meaningful alignment, enabling CaReFlow to achieve good transformation effect without recursively training the rectified flow model for multiple iterations \cite{liuflow,lee2024improving} and addressing the ambiguity problem in one-to-many mapping \cite{guo2025variational}. Secondly, to prevent the loss of discriminative information from the source modality during distribution transformation, we employ a `\textbf{cyclic rectified flow}' objective. This cycle-consistency constraint ensures features output by forward rectified flow can be mapped back to their original features \cite{MCTN}, guaranteeing that the transformed features can retain and interpret sufficient modality-specific information for fusion.

\textbf{Our contributions are summarized as follows:}
\begin{itemize}
    \item We reformulate the modality gap problem in MAC as a distribution mapping task and adapt rectified flow to solve it for the first time, enabling a more effective fusion.
    \item We propose a novel CaReFlow featuring one-to-many mapping, adaptive relaxed alignment, and cyclic consistency to enable a faster, accurate, robust and information-preserving learning of modality transformation.
    \item We demonstrate that after reducing modality gap using our method, even simple fusion methods achieve state-of-the-art results on multiple MAC benchmarks. Visualizations further provide direct evidence that CaReFlow effectively narrows the modality gap in the feature space.
\end{itemize}

\section{Related Work}
\label{sec:related}

\subsection{Multimodal Affective Computing}
Previous algorithms for MAC mostly focus on proposing advanced fusion techniques such as tensor fusion \cite{Zadeh2017Tensor,koromilas2023mmatr_tensor} and attention-based fusion \cite{10224356,pramanick2022multimodal} to generate discriminative multimodal representations \cite{Zadeh2018Memory,yang2025towards,huang2025atcaf}.
In addition, many methods utilize Kullback–Leibler divergence, canonical correlation analysis and information bottleneck to regularize unimodal distributions  \cite{nll,MIB,luo2025triagedmsa}. For example, information-theoretic hierarchical perception (ITHP) \cite{ithp} applies information bottleneck to learn features, identifying a primary modality and treating other modalities as detectors that distill the flow of information.
Recently, multimodal large language models have emerged, leveraging large pre-trained models to interpret human affective states effectively \cite{zhao2025humanomnilargevisionspeechlanguage,cheng2024videollama,xu2025qwen2}.
Moreover, uncertainty-based fusion methods  \cite{gao2024embracing, ma2023calibrating,xu2024reliable,10287630} improve the reliability of fusion by estimating the predictive confidence of each modality. 

Those deep learning models focus on reducing modality gap are related to CaReFlow. Early methods generally employ techniques such as recurrent neural networks, cross-modal transformer and GANs to translate source modality to target modality \cite{MCTN,MULT,ARGF,nips_cca}. Later approaches utilize powerful diffusion models to achieve transformation between modality distributions, but they are slow in inference \cite{imder,lee2025diffusion}. Moreover, the aforementioned methods typically achieve one-to-one modality mapping within each sample, without presenting a broader overall distribution of target modality to the data points of source modality. Moreover, they are limited by insufficient modality pairs with each sample. Recently, contrastive learning has been used to reduce the modality gap \cite{yang2023confede,selfcon1}. Early contrastive learning methods focus on constructing positive modality pairs within each sample, while later methods construct positive pairs within the same category, significantly increasing the number of training pairs \cite{hycon,yang2024clgsi}. However, contrastive learning does not differentiate the importance between modality pairs from the same sample and those from different samples, failing to achieve more accurate modality alignment. In contrast, drawing inspiration from the simple yet effective rectified flow \cite{liuflow,li2025omniflow,guo2025variational}, CaReFlow enables one-to-many mapping for more effective alignment, and implements adaptive relaxed alignment for different modality pairs with distinct correlation degrees to achieve faster and more accurate alignment. Moreover, CaReFlow realizes cyclic information flow to avoid information loss.

\subsection{Rectified Flow}

Rectified flow is a powerful tool that can map one distribution to another by learning a straight and fast trajectory \cite{ma2025janusflow,liuflow}, which has been applied in various generation tasks \cite{li2025omniflow}. It iteratively learns smooth ordinary differential equation (ODE) paths that are less susceptible to truncation error, but the recursive training with multiple iterations suffers from high computational cost. Lee \textit{et al.} \cite{lee2024improving} show that under realistic settings, a single iteration for training rectified flow with specifically designed techniques is sufficient to learn nearly straight trajectories. 
Guo \textit{et al.} \cite{guo2025variational} further propose a variational rectified flow matching that introduces a latent variable to disentangle ambiguous flow directions at each location, which addresses the ambiguity problem in one-to-many mapping.
In contrast, we extend rectified flow to reduce the modality gap for fusion. We show that we can simultaneously avoid recursive training and address the ambiguous problem in one-to-many mapping by the proposed adaptive relaxed alignment. It imposes stricter alignment constraints on modality pairs with closer relationships and applies more relaxed constraints on modality pairs that do not belong to the same category, enabling a more accurate and faster learning of rectified flow. Moreover, we propose cyclic information flow that ensures the latent features output by rectified flow can be transferred back to the original features, avoiding information loss of the source modality.  
\section{Methodology}
\begin{figure*}[t]
\centering
\includegraphics[width=0.75\linewidth]{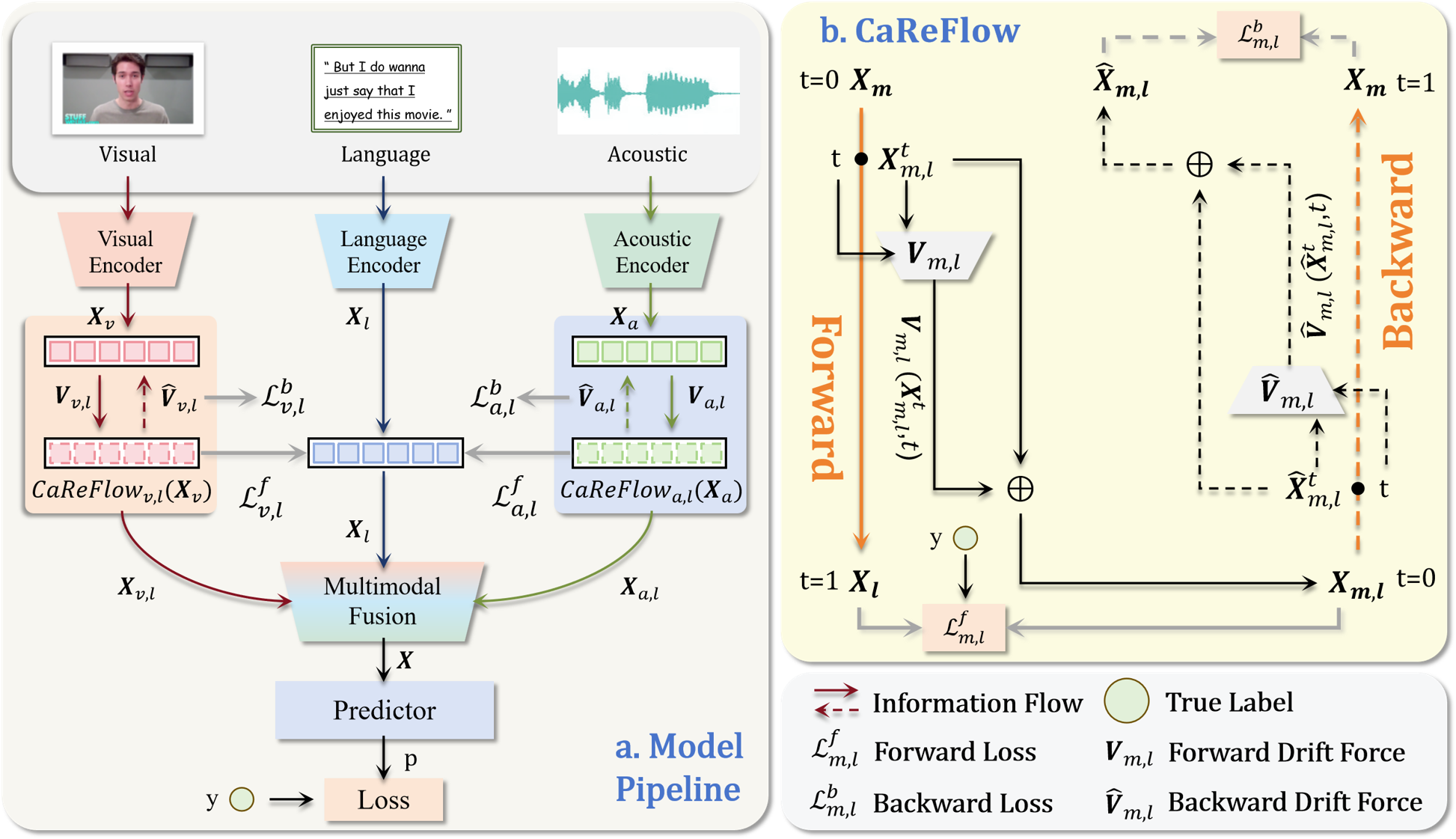}
\vspace{-0.2cm}
\caption{(a) The diagram of the model pipeline and (b) the forward and backward information flows in CaReFlow.}
\label{fig:framework}
\vspace{-0.3cm}
\end{figure*}

The pipeline of CaReFlow is shown in Figure~\ref{fig:framework}, which maps the distributions of visual and acoustic modalities to language modality using cyclic adaptive rectified flow and then conducts fusion for prediction. CaReFlow is evaluated on multiple MAC tasks, including multimodal sentiment analysis (MSA) \cite{zadeh2016multimodal}, multimodal humor detection (MHD) \cite{ur_funny} and multimodal sarcasm detection (MSD) \cite{msd}.

\subsection{Preliminaries} \label{sec:pre}
(1) \textbf{MAC} aims to predict the sentiment scores, affective states, opinion tendency, and behavior intentions given a video segment of a speaker described by acoustic ($a$), visual ($v$), and language ($l$) modalities. The input feature sequences are denoted as $\{\bm{U}_{m}\! \in\! \mathbb{R}^{T_m \times d_m } | m\! \in\! \{a, v, l\} \}$, where $T_m$  is the sequence length and $d_m$ is the feature dimensionality.
Unimodal networks are utilized to extract unimodal representations $\{\bm{X}_{m} \in \mathbb{R}^{d }\ |\ m \in \{a, v, l\} \}$  based on input feature sequences $\bm{U}_{m}$, where $d$ is the shared latent feature dimensionality. Due to space limitation, unimodal networks are introduced in the Appendix.

(2) \textbf{Rectified flow} is a simple and effective generative model that smoothly transitions between two distributions $p_{m_1}$ and $p_{m_2}$ by solving ordinary differential equations (ODEs) \cite{liuflow}. In the context of MAC, given $\bm{X}_{m_1} \sim p_{m_1}$ and $\bm{X}_{m_2} \sim p_{m_2}$, the interpolation between $\bm{X}_{m_1}$ and $\bm{X}_{m_2}$ is defined as $\bm{X}_{m_1, m_2}^t = (1-t)\cdot \bm{X}_{m_1} + t \cdot \bm{X}_{m_2}$ for $t \in [0, 1]$, where $\bm{X}_{m_1}$ and $\bm{X}_{m_2}$ are unimodal features for source and target modalities, respectively. The rectified flow induced from ($\bm{X}_{m_1}$, $\bm{X}_{m_2}$) is an ODE for time $t \in [0, 1]$:
\begin{equation}
\setlength{\abovedisplayskip}{2pt}
\setlength{\belowdisplayskip}{2pt}
\label{eq_ode}
d \bm{Z}^t_{m_1, m_2} = \bm{V}_{m_1, m_2}(\bm{Z}^t_{m_1, m_2}, t) dt
\end{equation}
which converts $\bm{Z}_{m_1}$ from source distribution $p_{m_1}$ to a $\bm{Z}_{m_2}$ in target distribution $p_{m_2}$ during inference ($\bm{Z}_{m_1}$ is a data point in inference stage). The drift force $\bm{V}_{m_1, m_2}: \mathbb{R}^d \longrightarrow \mathbb{R}^d $ (also known as the velocity vector-field), often realized by a neutral network, is expected to drive the flow to follow the direction ($\bm{X}_{m_1}$, $\bm{X}_{m_2}$) as much as possible via the generated velocity vector. To realize this goal, we can solve a simple least squares regression problem during training:
\begin{equation}
\setlength{\abovedisplayskip}{2pt}
\setlength{\belowdisplayskip}{2pt}
\label{eq_reflow}
\min_{\bm{V}_{m_1, m_2}}  \int_0^1\! \mathbb{E}\! \left[ \left\| \bm{V}_{m_1, m_2}(\bm{X}_{m_1, m_2}^t, t)\! -\! (\bm{X}_{m_2}\!\! -\! \bm{X}_{m_1}) \right\|^2\! dt \right]
\quad 
\end{equation}
where $\bm{X}_{m_1, m_2}^t\! =\! (1\!-\!t)\cdot \bm{X}_{m_1}\! +\! t \cdot \bm{X}_{m_2}$ is the interpolation between $\bm{X}_{m_1}$ and $\bm{X}_{m_2}$. The objective in Eq.~\ref{eq_reflow} makes $\bm{X}_{m_1}+\bm{V}_{m_1, m_2}(\bm{X}_{m_1, m_2}^t, t)$ approximate $\bm{X}_{m_2}$ as closely as possible.
During training, rectified flow randomly selects data pairs from two distributions and then uses Eq.~\ref{eq_reflow} to train $\bm{V}_{m_1, m_2}$ via stochastic gradient descent. When $\bm{V}_{m_1, m_2}$ is well trained, using the learned drift force model $\bm{V}_{m_1, m_2}$ defined at time $t$ and location $\bm{Z}_{m_1, m_2}^t$, we can solve the ODE starting from $\bm{Z}_{m_1} \sim p_{m_1}$ (time $t=0$) to transfer source distribution $p_{m_1}$ to target distribution $p_{m_2}$  (time $t=1$), which is often simulated with Euler steps during inference. 

Applying the rectified flow operation recursively (using the outputs of previous rectified flow model to train new rectified flow model) yields a sequence of rectified flow models, which can straighten paths of rectified flows (make the paths of the flow more straight). 
Perfectly straight paths can be simulated exactly with a single Euler step during inference ($\bm{Z}_{m_1, m_2} = \bm{Z}_{m_1} + \bm{V}_{m_1, m_2}(\bm{Z}_{m_1}, 1)$ with $t$ being 1) and is effectively a one step model. 
Nevertheless, the proposed CaReFlow can avoid this recursive operation and generate desirable $\bm{Z}_{m_1, m_2}$ using two Euler steps without training the rectified flow model for multiple times, which significantly reduces computational cost.
In the remainder, for the sake of consistency, we will no longer distinguish between $\bm{X}$ and $\bm{Z}$, and uniformly use $\bm{X}$ instead.

\subsection{Model Pipeline} \label{sec:pipeline}
In this paper, we focus on adapting rectified flow to narrow the gap between different modalities in the context of multimodal fusion. Given unimodal representations $\bm{X}_{m}$, as language is the dominant modality in MAC \cite{mai2025injecting,MCTN}, we map the distributions of visual and acoustic modalities to that of the language modality using CaReFlow, and then conduct effective multimodal fusion and prediction:
\begin{equation}
\setlength{\abovedisplayskip}{3pt}
\setlength{\belowdisplayskip}{3pt}
\label{eq_CaReFlow1}
\begin{split}
\bm{X}_{m, l} = CaReFlow_{m,l}(\bm{X}_{m}), \ \ m \in \{a, v \}
 \end{split}
\end{equation}
\begin{equation}
\setlength{\abovedisplayskip}{3pt}
\setlength{\belowdisplayskip}{3pt}
\label{fusion}
 \bm{X} =  Fusion(\bm{X}_{l}, \bm{X}_{a, l}, \bm{X}_{v, l}; \ \theta_f)
\end{equation}
\begin{equation}
\setlength{\abovedisplayskip}{3pt}
\setlength{\belowdisplayskip}{3pt}
\label{predictor}
 p =  Predictor(\bm{\bar{X}}; \theta_{pre});\ \ \ \mathcal{L} = Loss(p,y)
\end{equation}
where $\bm{X}$ is the multimodal representation, $p$ is the final prediction, and $\mathcal{L}$ is the predictive loss (the concrete form of the loss function $Loss$ depends on downstream tasks).
In practice, we simply concatenate the unimodal features and then use a multi-layer perception (MLP) network to construct the fusion network $Fusion$. We show that even with such simple fusion network, the proposed method still obtains very competitive performance after distribution mapping.

In particular, we apply two Euler steps to generate $\bm{X}_{m, l}$, in which case $d t$ in Eq.~\ref{eq_ode} is equal to 0.5. Thus, $CaReFlow_{m,l}(\bm{X}_{m})$ in Eq.~\ref{eq_CaReFlow1} is defined as:
\begin{equation}
\setlength{\abovedisplayskip}{3pt}
\setlength{\belowdisplayskip}{3pt}
\label{eq_CaReFlow2}
\begin{split}
  & \bm{X}_{m, l}^{t_{=0.5}} = \bm{X}_{m} + \bm{V}_{m, l}(\bm{X}_{m}^{t_{=0}}, 0) \times 0.5 \\
 \bm{X} & _{m, l} = \bm{X}_{m, l}^{t_{=0.5}} + \bm{V}_{m, l}(\bm{X}_{m, l}^{t_{=0.5}}, 0.5) \times 0.5
 \end{split}
\end{equation}
It is obvious that $\bm{X}_{m, l}^{t_{=0}}$ is equal to $\bm{X}_{m}$. It can also be inferred from Eq.~\ref{eq_CaReFlow2} that the features of target modality $\bm{X}_{l}$ is not involved in the inference stage of rectified flow, leading to a causal information flow from source to target modality.

\subsection{Cyclic Adaptive Rectified Flow}

Here we introduce CaReFlow in detail, which extends the regular rectified flow with cyclic latent information flow and adaptive relaxed mapping.  Notably, CaReFlow is operated in the latent feature space, which significantly reduces running time and enables an easier distribution mapping.

\subsubsection{Adaptive Relaxed Modality Alignment}
As stated in Section~\ref{sec:pre}, rectified flow randomly selects many data pairs from two distributions to train $\bm{V}_{m_1, m_2}$ (see Eq.~\ref{eq_reflow}), which enables the data points of the source modality to access the global distribution information of the target modality. However, rectified flow equally treats all pairs during mapping, which leads to \textbf{ambiguous one-to-many mapping} for each data point in the source modality because it is required to match many data points in the target modality \cite{guo2025variational}. 
Actually, there exists accurate one-to-one mapping relationship between modalities from the same sample. 
Intuitively, we should strictly align modality pairs from the same sample, while applying relaxed alignment for pairs from different samples. Moreover, considering that samples from the same category tend to have more similar distributions, we should impose stricter alignment on modality pairs from the same category than those from different categories. In this way, CaReFlow not only allows the data points from source modality to gain a global understanding of the distribution of target modality, but also achieves more accurate distribution mapping. It effectively addresses the ambiguous one-to-many mapping problem by guiding the model to recognize which data points in the target modality should be focused more on during mapping. 
Moreover, by performing more accurate alignment, adaptive relaxed alignment enables CaReFlow to achieve satisfactory alignment effect without iteratively training the rectified flow model in cycles, achieving a faster training.

Specifically, by incorporating adaptive relaxed alignment, the objective of CaReFlow becomes:
\begin{equation}
\setlength{\abovedisplayskip}{2pt}
\setlength{\belowdisplayskip}{2pt}
\label{eq_careflow}
\begin{split}
\mathcal{L}^f_{m_1, m_2}\!  =\! \small \mathbb{E}\! & \ \ \Big[ \ \big\| max(\bm{V}_{m_1, m_2}(\bm{X}_{m_1, m_2}^t, t)\! \\ &  -\! (\bm{X}_{m_2}\!  -\! \bm{X}_{m_1}) \big\|^2\!  - \eta_{m_1, m_2}, 0) \ \Big]  \\
s.t. \ \ \bm{X}_{m_1} & \sim p_{m_1},\ \bm{X}_{m_2} \sim p_{m_2},\ t \in [0, 1]
\end{split}
\end{equation}
where $\mathcal{L}^f_{m_1, m_2}$ is the loss for forward rectified flow, $\eta_{m_1, m_2}$ is the margin that controls how `strict' the modality mapping should be. Particularly, according to whether $\bm{X}_{m_1}$ and $\bm{X}_{m_2}$ belong to the same sample, $\eta_{m_1, m_2}$ is computed as:
\begin{equation}
\setlength{\abovedisplayskip}{4pt}
\setlength{\belowdisplayskip}{4pt}
\begin{aligned}
   \eta_{m_1, m_2}\! =\!
   \begin{cases}
   0, &  \text{from the same sample} \\
   \epsilon \! +\! || y_i - y_j ||^2 , & \text{from different samples} 
   \end{cases}
\end{aligned}
\end{equation}
where $\epsilon$ is a hyperparameter, and $y_i$ denotes the ground-truth label of sample $i$. Assuming $\bm{X}_{m_1}$ and $\bm{X}_{m_2}$ come from sample $i$ and $j$ respectively, $|| y_i - y_j ||^2$ denotes the label distance between two data points. If $\bm{X}_{m_1}$ and $\bm{X}_{m_2}$ come from the same sample, $\eta_{m_1, m_2}$ is zero and the training objective in Eq.~\ref{eq_careflow} reduces to a regular training objective in Eq.~\ref{eq_reflow}. And if $\bm{X}_{m_1}$ and $\bm{X}_{m_2}$ come from different samples but sharing the same category, $\eta_{m_1, m_2}$ is equal to $\epsilon$, and the forward loss $\mathcal{L}^f_{m_1, m_2}$ becomes zero if $||(\bm{X}_{m_1}+\bm{V}_{m_1, m_2}(\bm{X}_{m_1, m_2}^t, t)) - \bm{X}_{m_2}||^2$ is less than $\epsilon$, which leads to relaxed mapping. Finally, if $\bm{X}_{m_1}$ and $\bm{X}_{m_2}$ come from different categories ($|| y_i - y_j ||^2$ is defined as 1 for classification tasks), $\eta_{m_1, m_2}$ is larger than $\epsilon$, resulting in a more relaxed alignment. In this way, we can learn modality mapping relationships in a faster and more accurate way without training rectified flow model for multiple times. 

In practice, for each mini-batch, we first construct modality pairs within each sample that have accurate correspondences between the two modalities. Then, we randomly sample multiple modality pairs between different samples to train the model, enabling the data points from the source modality to access the overall distribution information of the target modality. In this process, we set the number of modality pairs from different samples to be $\beta$ times the number of modality pairs from the same sample.

\textbf{The Implementation of $\bm{V}_{m_1, m_2}$}: As stated in Section~\ref{sec:pre}, $\bm{V}_{m_1, m_2}$ is parameterized as a neural network. Since $\bm{V}_{m_1, m_2}$ takes both time-dependent representation $\bm{X}_{m_1, m_2}^t$ and time $t$ as inputs, it is important to implement $\bm{V}_{m_1, m_2}$ in a time-dependent manner to generate time-aware output. In practice, we follow the positional embedding of Transformer \cite{transformer} to use sine and cosine functions of different frequencies to reveal the time information, which does not introduce learnable parameters:
\begin{equation}
\setlength{\abovedisplayskip}{3pt}
\setlength{\belowdisplayskip}{3pt}
\label{eq_time}
\begin{split}
 \bm{TE}_{2i}^t = sin(t * 1000/10000^{2i/d}) & \\
 \bm{TE}_{2i + 1}^t = cos(t * 1000/10000^ {2i/ d }&)
\end{split}
\end{equation}
where $\bm{TE}^t\in \mathbb{R}^d $ is the time embedding and $i$ is the dimension. After obtaining  $\bm{TE}^t$, we directly concatenate input features with the time embedding, and use a simple multi-layer perception (MLP) network to generate the output:
\begin{equation}
\setlength{\abovedisplayskip}{3pt}
\setlength{\belowdisplayskip}{3pt}
\label{eq_v}
\bm{V}_{m_1, m_2}(\bm{X}_{m_1, m_2}^t, t)\! =\! MLP(\bm{X}_{m_1, m_2}^t \otimes \bm{TE}^t)
\end{equation}
where $\otimes$ denotes the concatenation operation.
It is worth noting that we apply the `detach' operation to the input features of $\bm{V}_{m_1, m_2}$ to prevent the forward loss $\mathcal{L}_f$ from updating unimodal networks. In other words, the goal of $\mathcal{L}_f$ is to achieve distribution transformation between different modalities, rather than training better unimodal networks. 

\subsubsection{Cyclic Latent Information Flow}

Although the transition from $p_{m_1}$ to $p_{m_2}$ is relatively simple in rectified flow, it might still lead to information loss of source modality 
during modality mapping. To prevent information loss, we construct a \textbf{cyclic rectified flow} that maps the output of the forward rectified flow $\bm{X}_{m_1, m_2}$ to the original features $\bm{X}_{m_1}$. In this way, we can ensure that $\bm{X}_{m_1, m_2}$ can retain and interpret the information from the source modality, such that the obtained multimodal representation $\bm{X}$ can learn more sufficient modality-specific information for prediction. 
Notably, here we only construct modality pairs within the same sample to ensure that the latent features output by forward rectified flow can be transformed back to the original modality features. Therefore,
the objective of the backward rectified flow is defined as:
\begin{equation}
\setlength{\abovedisplayskip}{2pt}
\setlength{\belowdisplayskip}{2pt}
\begin{split}
\label{eq_careflow_b}
 \mathcal{L}^b_{m_1, m_2} &\! =\! \mathbb{E}\! \ \Big[ \ \big\| \bm{\hat{V}}_{m_1, m_2}(\bm{\hat{X}}_{m_1, m_2}^t, t)\! \\ &-\! (\bm{X}_{m_1}\! -\! \bm{X}_{m_1, m_2}) \big\|^2 \ \Big]  
\quad \\
s.t. \ \bm{X}_{m_1} \sim  \ & p_{m_1},\ \bm{X}_{m_1, m_2} \sim p_{m_1, m_2},\ t \in [0, 1]
\end{split}
\end{equation}
where $\bm{\hat{X}}_{m_1, m_2}^t = (1-t)\cdot \bm{X}_{m_1, m_2} + t \cdot \bm{X}_{m_1}$, $\bm{X}_{m_1, m_2}$ is the latent features generated by the forward rectified flow, $\bm{\hat{V}}_{m_1, m_2}$ is the drift force neural network for backward rectified flow, and $\mathcal{L}^b_{m_1, m_2}$ denotes the backward loss. 
Here we only apply the detach operation to $\bm{X}_{m_1}$, but not to $\bm{X}_{m_1, m_2}$, allowing $\mathcal{L}^b_{m_1, m_2}$ to influence the forward rectified flow, thereby aiding in learning a better distribution transformation that preserves modality-specific information.

\subsection{Model Optimization}
The optimization target of CaReFlow is the combination of main task loss, forward loss, and backward loss:
\begin{equation}
\setlength{\abovedisplayskip}{3pt}
\setlength{\belowdisplayskip}{3pt}
\label{eq122}
 \mathcal{L}_{total} = \mathcal{L} + \sum_{m}^{m \in \{a, v \}} ( \alpha_f \times \mathcal{L}^f_{m,l} + \alpha_b \times \mathcal{L}^b_{m,l})
\end{equation}
where $\alpha_f$ and $\alpha_b$ are hyperparameters for the weights of forward and backward losses, respectively. 
\section{Experiments}
\label{sec:experiments}
CaReFlow is evaluated on CMU-MOSI \cite{zadeh2016multimodal}, CMU-MOSEI \cite{MOSEI}, CH-SIMS-v2 \cite{simsv2}, UR-FUNNY \cite{ur_funny}, and MUStARD \cite{msd} datasets. Due to space limitation, \textbf{we introduce experimental settings, baselines, datasets and additional experimental results in the Appendix.}

\subsection{Results on the MSA Task}

The results on the MSA task are shown in Table~\ref{tbase} and Table~\ref{tab:SIMSResult}.
Specifically, on CMU-MOSI, CaReFlow outperforms the state-of-the-art baseline DLF \cite{wang2025dlf} by over 1 points in Acc7 and Acc2. 
On CMU-MOSEI, CaReFlow obtains the best performance in Acc2, F1 score, MAE and Corr, and the second highest score in Acc7. 
Similar results are observed on the CH-SIMS-v2 dataset (see Table~\ref{tab:SIMSResult}), where CaReFlow considerably outperforms all baselines in all evaluation metrics (over 4 points improvement in Acc5).
Generally, \textbf{CaReFlow demonstrates state-of-the-art results on MSA across three widely-used datasets}. This is mainly because CaReFlow effectively addresses modality gap via the cyclic adaptive rectified flow before multimodal fusion, which can improve the effectiveness of fusion.

\begin{table*}
\centering
\small
 \vspace{-0.2cm}
 \caption{ \label{tbase}The results on CMU-MOSI and CMU-MOSEI. The results labeled with $^{\dag}$ are obtained from original papers, and other results are obtained from our experiments. The best results are highlighted in bold and the second best results are marked with underlines.
 }
 \vspace{-0.1cm}
 \resizebox{2.0\columnwidth}{!}{\begin{tabular}{c||c|c|c|c|c|c|c|c|c|c|c}
  \noalign{\hrule height 1pt} 
\rowcolor{lightgray!40} 
   &  & \multicolumn{5}{c|}{CMU-MOSI} & \multicolumn{5}{c}{CMU-MOSEI}  \\
 \cline{3-12}
 \rowcolor{lightgray!40}
\multirow{-2}{*}{Model} & \multirow{-2}{*}{Venue} & Acc7$\uparrow$  & Acc2$\uparrow$ & F1$\uparrow$ & MAE$\downarrow$ & Corr$\uparrow$ & Acc7$\uparrow$  & Acc2$\uparrow$ & F1$\uparrow$ & MAE$\downarrow$ & Corr$\uparrow$\\
\hline
\hline
    Self-MM \citep{mmsa} & AAAI 2021 &  45.8 &   84.9  &   84.8  & 0.731  & 0.785  &  53.0 &   85.2  &   85.2  &  0.540 & 0.763 \\
    C-MIB \citep{MIB} & TMM 2023 &  47.7 &   87.8 &  87.8  & 0.662  & 0.835  & 52.7 &  86.9  & 86.8 & 0.542  &  0.784\\
    DMD \citep{li2023decoupled_cvpr} & CVPR 2023 & 44.9 & 84.3  &   84.3 & 0.726 & 0.788  &   52.8 &  84.6  &  84.6  & 0.538 & 0.768 \\
    DEVA$^{\dag}$ \citep{deva} & AAAI 2025 & 46.3 &  86.3 &  86.3  &  0.730 &  0.787 & 52.3 &  86.1   &  86.2  &  0.541 &  0.769 \\
AtCAF$^{\dag}$ \citep{huang2025atcaf} & Information Fusion 2025  & 46.5  & 88.6  &  88.5  &  0.650 & 0.831  & \textbf{55.9}  & 87.0  &  86.8 & \underline{0.508} & 0.785 \\
EMOE \citep{EMOE2025} & CVPR 2025 & 45.2  & 84.8  &  84.8  &  0.723 & 0.790  & 52.5  & 85.0  &  85.0 & 0.542 & 0.760 \\
Multimodal Boosting \citep{10224356} & TMM 2024 & 49.1  &  88.5 & 88.4  &  \underline{0.634} & 0.855  & 54.0  &  86.5 & 86.5 & 0.523  & 0.779 \\
ITHP \citep{ithp} & ICLR 2024 &  47.7 &   88.5 &   88.5  & 0.663  & \underline{0.856}  & 52.2  &   87.1  & 87.1 &  0.550 & \underline{0.792} \\
DLF \citep{wang2025dlf} & AAAI 2025 & \underline{49.4}  & \underline{88.7}  &  \underline{88.6}  &  0.669 & \underline{0.856}  & 55.0  & \underline{87.5}  &  \underline{87.4} & 0.515 & 0.774 \\
    \hline
    \rowcolor[HTML]{EBFAFF}
 CaReFlow & -  & \textbf{50.6}  &   \textbf{89.8} &  \textbf{89.7}  &  \textbf{0.616} & \textbf{0.858}  & \underline{55.7} &  \textbf{87.9} &   \textbf{88.0} & \textbf{0.504}  &  \textbf{0.799} \\
    \noalign{\hrule height 1pt} 
     \end{tabular}}
     \vspace{-0.2cm}
\end{table*}%

\begin{table}[t]
    \vspace{0.1cm}
    \caption{The comparison with baselines on CH-SIMS-v2. 
    }
    \vspace{-0.1cm}
    \label{tab:SIMSResult}
    \resizebox{0.48\textwidth}{!}{
        \begin{tabular}{c||c|c|c|c|c|c}
        \noalign{\hrule height 1pt} 
\rowcolor{lightgray!40}
         & \multicolumn{6}{c}{CH-SIMS v2} \\ 
        \cline{2-7}
        \rowcolor{lightgray!40}
       \multirow{-2}{*}{Model}  & Acc5↑ & Acc3↑ & Acc2↑ & F1↑ & MAE↓ & Corr↑  \\ \hline
       \hline
        MISA \cite{MISA} & 47.5 & 68.9 & 78.2 & 78.3 & 0.342 & 0.671 \\
        MAG-BERT \cite{MAG-BERT} & 49.2 & 70.6 & 77.1 & 77.1 & 0.346 & 0.641 \\
        Self-MM \cite{mmsa} & \underline{53.5} & 72.7 & 78.7 & 78.6 & 0.315 & 0.691 \\
        MMIM \cite{MMIM} & 50.5 & 70.4 & 77.8 & 77.8 & 0.339 & 0.641 \\
        AV-MC \cite{simsv2} & 52.1 & 73.2 & \underline{80.6} & \underline{80.7} & 0.301 & 0.721 \\
        KuDA \cite{kuda} & 53.1 & \underline{74.3} & 80.2 & 80.1 & \underline{0.289} & \underline{0.741} \\
        DLF \citep{wang2025dlf} & 47.5 & 70.0 & 78.1 & 77.9 & 0.346 & 0.683  \\  \hline
        \rowcolor[HTML]{EBFAFF}
        CaReFlow & \textbf{57.9} & \textbf{75.9} & \textbf{82.9} & \textbf{82.9} & \textbf{0.277} & \textbf{0.745}  \\  
         \noalign{\hrule height 1pt} 
        \end{tabular}
    }
    \vspace{-0.2cm}
\end{table}

\subsection{Results on the MHD and MSD Tasks}



\begin{figure}
    \centering
    \begin{subfigure}[t]{0.5\linewidth} 
        \centering
        \includegraphics[width=\linewidth]{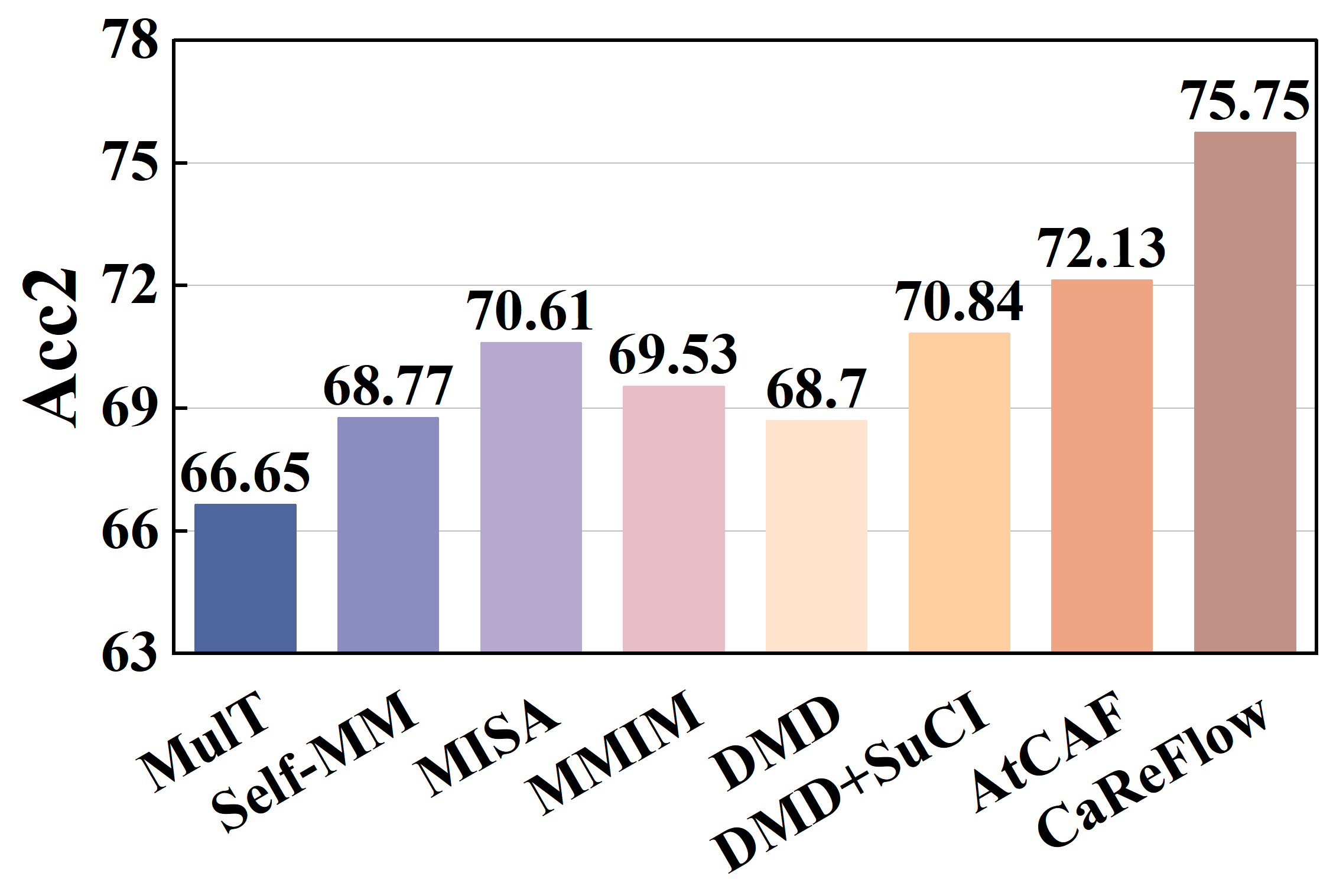}
        \vspace{-0.4cm}
        \caption{Results on MHD task}
    \end{subfigure}
    \begin{subfigure}[t]{0.48\linewidth}
        \centering
        \includegraphics[width=\linewidth]{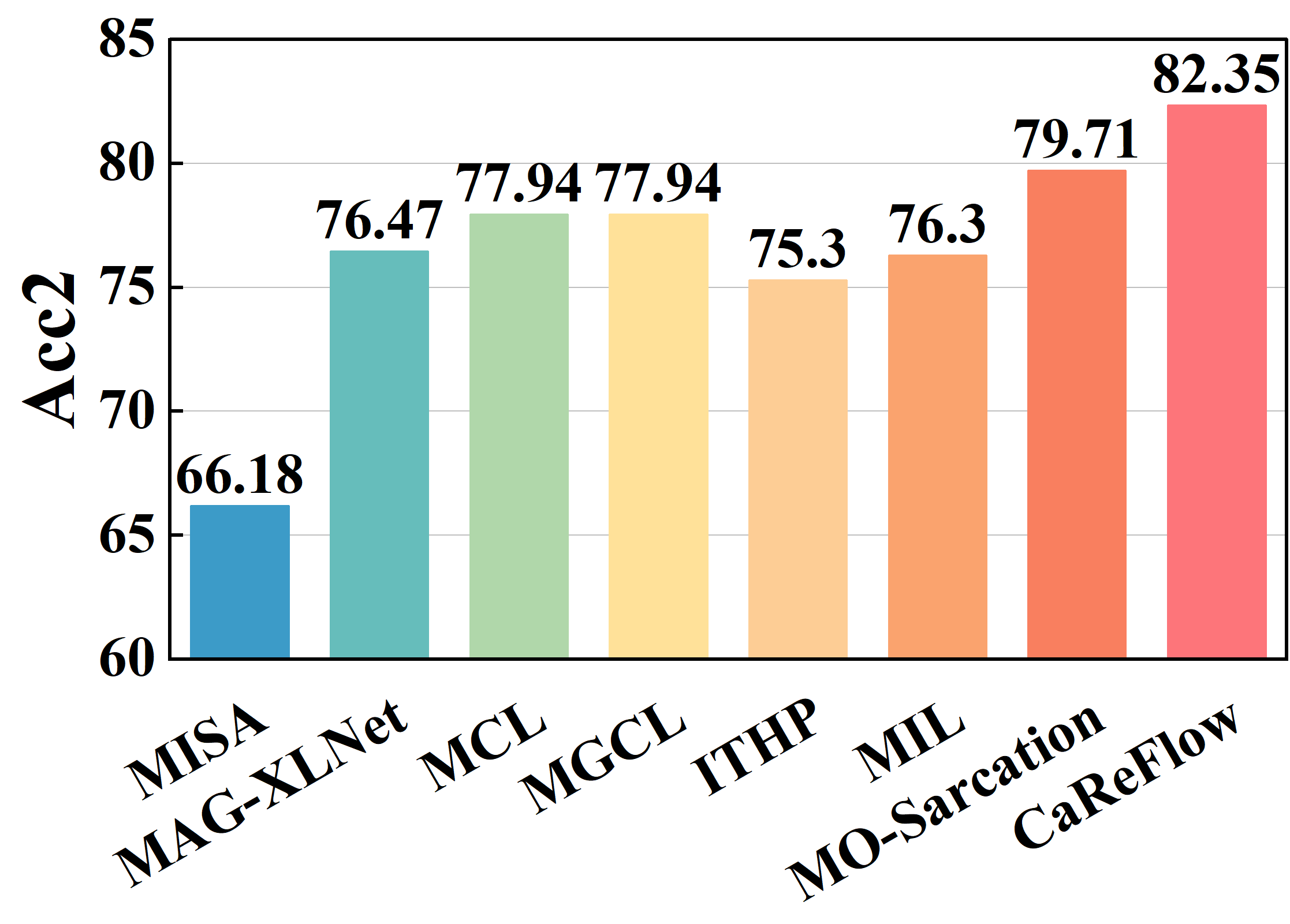}
        \vspace{-0.4cm}
        \caption{Results on MSD task}
    \end{subfigure}
    \vspace{-0.2cm}
    \caption{Results on (a) UR-FUNNY and (b) MUStARD datasets.}
    \label{mhd_msd}
    \vspace{-0.3cm}
\end{figure}

To evaluate the generalizability of CaReFlow to other multimodal tasks, we conduct experiments on MHD and MSD  using UR-FUNNY \cite{ur_funny} and MUStARD \cite{msd} datasets. The compared baselines include MulT \cite{MULT}, Self-MM \cite{mmsa}, MMIM \cite{MMIM}, DMD \cite{li2023decoupled_cvpr}, DMD+SuCI \cite{yang2024towards}, AtCAF \cite{huang2025atcaf}, MAG-XLNet \cite{MAG-BERT}, MCL \cite{mcl}, MGCL \cite{mgcl}, ITHP \cite{ithp}, MISA \cite{MISA}, MIL \cite{MIL}, and MO-Sarcation \citep{tomar2023your}.
As presented in Figure~\ref{mhd_msd}, CaReFlow outperforms the best performing baselines AtCAF and MO-Sarcation by over 3 points and 2.5 points on UR-FUNNY and MUStARD, respectively. 
Generally, \textbf{CaReFlow reaches state-of-the-art results on the MHD and MSD tasks}, demonstrating the effectiveness and \textbf{generalizability of CaReFlow to other tasks}.

\subsection{Ablation and Comparison Experiments}



\begin{table}
\vspace{-0.cm}
\centering
\caption{\label{t3}Ablation experiments on CMU-MOSI and CH-SIMS-v2.}
\vspace{-0.2cm}
\resizebox{\linewidth}{!}{
\setlength\tabcolsep{5pt}
\renewcommand\arraystretch{1.2}
\begin{tabular}{c||c|c|c|c|c|c}
\noalign{\hrule height 1pt} 
\rowcolor{lightgray!40}
& \multicolumn{3}{c|}{CH-SIMS-v2} & \multicolumn{3}{c}{CMU-MOSI} \\ 
\cline{2-7}
\rowcolor{lightgray!40}
\multirow{-2}{*}{\textbf{Model}}
& \textbf{Acc5}$\uparrow$  & \textbf{Acc2}$\uparrow$  & \textbf{MAE}$\downarrow$ & \textbf{Acc7}$\uparrow$  & \textbf{Acc2}$\uparrow$  & \textbf{MAE}$\downarrow$  \\ 
\hline \hline
  W/O Distribution Alignment & 55.1  & 78.4  & 0.312  & 45.9 & 86.7  &  0.661 \\
  W/O Cyclic Information Flow & 56.3  &  81.7  & 0.296  & 47.2 & 87.9 & 0.643 \\
  W/O Adaptive Relaxed Alignment & 56.8  & \underline{82.4}  & \underline{0.283}  & \underline{47.9}  & \underline{88.5} & \underline{0.639} \\
  W/O One-to-Many Mapping & \underline{57.4}  & 79.8  & 0.287  & 47.2  &  87.0 & 0.643\\
  \hline
\rowcolor[HTML]{EBFAFF}
CaReFlow & \textbf{57.9} & \textbf{82.9}  & \textbf{0.277}   & \textbf{50.6}  &  \textbf{89.8} & \textbf{0.616} \\  

\noalign{\hrule height 1pt} 
\end{tabular}
}
\vspace{-0.3cm}
\end{table}


\textbf{(1) Distribution Alignment}:
In `W/O Cyclic Distribution Alignment', modality distribution mapping is removed and CaReFlow reduces to a regular multimodal model.
As shown in  Table~\ref{t3}, the performance of CaReFlow decreases by 4.5 points in Acc2 on CH-SIMS-v2 and 4.7 points in Acc7 on CMU-MOSI, indicating the importance of CaReFlow to align modality distributions for a more effective fusion; 
\textbf{(2) Cyclic Information Flow}:
In`W/O Cyclic Information Flow', we remove the backward rectified flow that transfers the output of forward rectified flow back to its original features,  and the performance declines considerably on both datasets,
revealing that it is important to apply cyclic information flow. This is because it helps to retain and interpret the information of source modality, enabling multimodal representations to learn sufficient modality-specific information for prediction; 
\textbf{(3) Adaptive Relaxed Alignment}:
In `W/O Adaptive Relaxed Alignment', the training objective in Eq.~\ref{eq_careflow} reduces to the regular objective of rectified flow in Eq.~\ref{eq_reflow}, and the performance exhibits a decline in all metrics. This is because adaptive relaxed alignment helps to learn more accurate modality alignment and can address the ambiguous problem in one-to-many mapping;
\textbf{(4) One-to-Many Mapping}:
In `W/O One-to-Many Mapping', we do not use modality pairs from different samples to train CaReFlow, and the performance significantly declines in all metrics (about 3 points in Acc2). Moreover, the performance drop is more significant than the drop when other modules are removed, indicating that one-to-many mapping is crucial for performance improvement. This is because it allows CaReFlow to sample abundant pairs for a more effective and robust learning of distribution mapping.

\begin{table*}[t]
\centering
\small
\caption{ \label{xxx}Discussion on different distribution mapping methods on the CMU-MOSI, CMU-MOSEI, and CH-SIMS-v2 datasets.}
\vspace{-0.2cm}
\resizebox{2.0\columnwidth}{!}{\begin{tabular}{c||c|c|c|c|c|c|c|c|c|c|c|c|c|c|c|c|c}
\noalign{\hrule height 1pt} 
\rowcolor{lightgray!40}
  &\multicolumn{6}{c|}{CMU-MOSI} & \multicolumn{5}{c|}{CMU-MOSEI} & \multicolumn{6}{c}{CH-SIMS-v2} \\
 \cline{2-18}
 \rowcolor{lightgray!40}
 \multirow{-2}{*}{ Model} & Acc7$\uparrow$ & Acc2$\uparrow$ & F1$\uparrow$ & MAE$\downarrow$ & Corr$\uparrow$ & Paramaters$\downarrow$  & Acc7$\uparrow$ & Acc2$\uparrow$ & F1$\uparrow$ & MAE$\downarrow$ & Corr$\uparrow$ & Acc5$\uparrow$ & Acc3$\uparrow$ & Acc2$\uparrow$ & F1$\uparrow$ &  MAE$\downarrow$ & Corr$\uparrow$ \\
\hline
\hline
ARGF \cite{ARGF}  & \underline{50.5} & 87.4 & 87.3 & \underline{0.623} & 0.846 & \textbf{184.43M} & \underline{53.2} & \underline{87.6}  & \underline{87.5} & \underline{0.524} & 0.798 & 51.8 & 71.0 & 78.4 & 78.7 & 0.314 & 0.677  \\
MulT \cite{MULT} & 40.7 & 88.1 & 88.1 & 0.755 & 0.821 & 185.51M & 51.7 & 86.5 & 86.6 & 0.583 & 0.794 & \underline{54.6} & \underline{74.2} & \underline{80.8} & \underline{80.7} & \underline{0.300} & \underline{0.738}  \\
Deep CCA \cite{andrew2013deep} & 43.8 & 87.2 & 87.2 & 0.722 & 0.815 & \underline{184.59M} & 51.0 & 84.6 & 84.7 & 0.574 & 0.775 & 51.3 & 71.7 & 77.9 & 77.9 & 0.334 & 0.668 \\
CLGSI \cite{yang2024clgsi} & 45.8 & \underline{89.0} & \underline{88.9} & 0.642 & \underline{0.851} & 186.31M & \textbf{55.7} & 87.2 & 87.1 & \textbf{0.504} & 0.797 & 52.5 & 71.0 & 78.2 & 78.3 & 0.313 & 0.683 \\
Diffusion Bridge \cite{lee2025diffusion} & 47.3 & 86.9 & 86.8 & 0.649 & 0.839 & 185.46M & 53.1 & 87.1 & 87.0 & 0.531 & \textbf{0.800} &  52.5 & 70.7 & 78.6 & 79.0 & 0.323 & 0.677 \\
\rowcolor[HTML]{EBFAFF}
\hline
CaReFlow & \textbf{50.6}  &   \textbf{89.8} &  \textbf{89.7}  &  \textbf{0.616} & \textbf{0.858} & 185.38M   & \textbf{55.7} &  \textbf{87.9} &   \textbf{88.0} & \textbf{0.504}  &  \underline{0.799} & \textbf{57.9} & \textbf{75.9} & \textbf{82.9} & \textbf{82.9} & \textbf{0.277} & \textbf{0.745} \\
\noalign{\hrule height 1pt} 
\end{tabular}}
\vspace{-0.1cm}
\end{table*}%

\begin{figure*}[t]
    \centering
    \begin{subfigure}[b]{0.19\linewidth} 
        \centering
        \includegraphics[width=\linewidth]{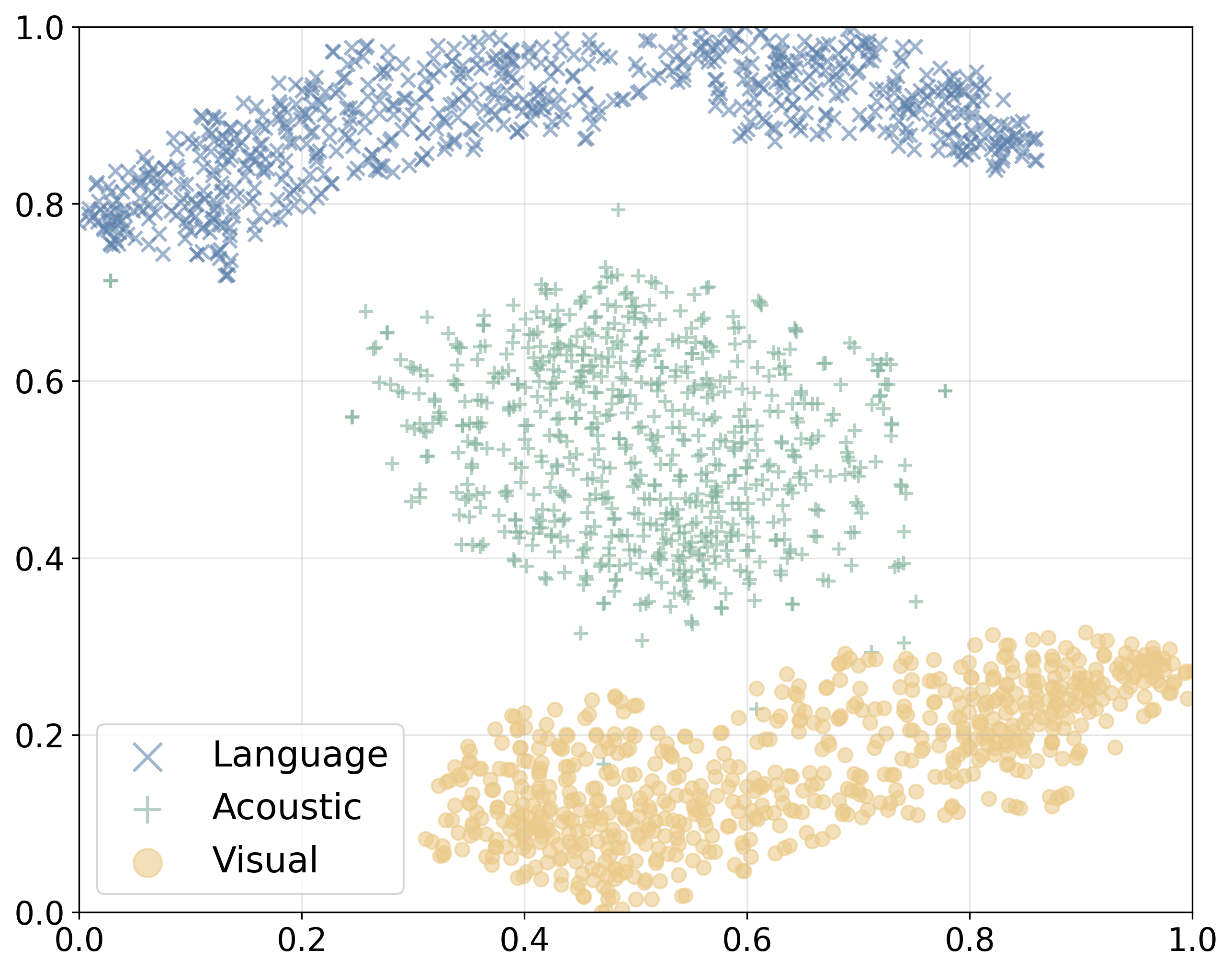}
        \caption{ARGF (GAN) }
    \end{subfigure}
    \begin{subfigure}[b]{0.19\linewidth}
        \centering
        \includegraphics[width=\linewidth]{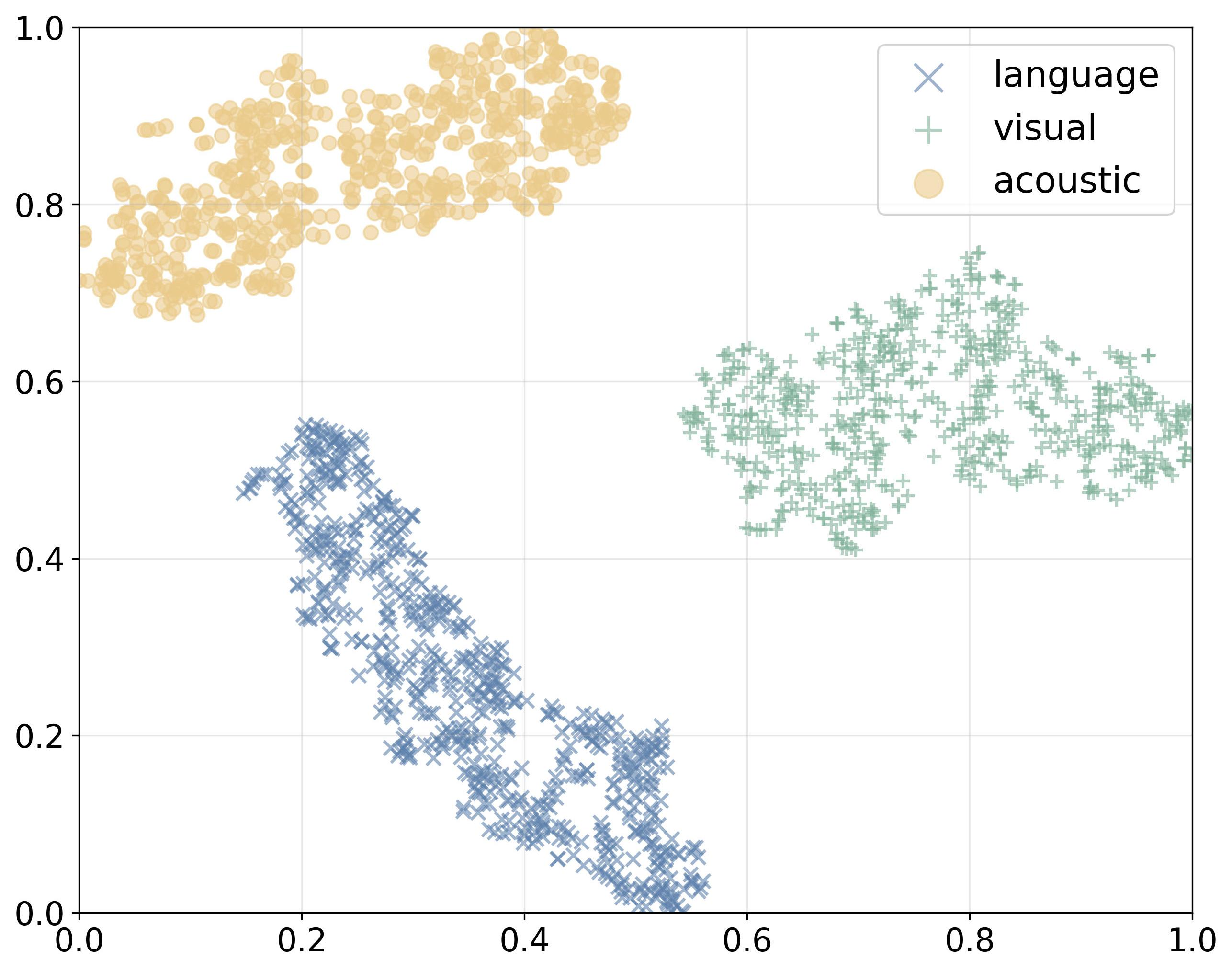}
        \caption{CLGSI (CL)  }
    \end{subfigure}
    \begin{subfigure}[b]{0.19\linewidth}
        \centering
        \includegraphics[width=\linewidth]{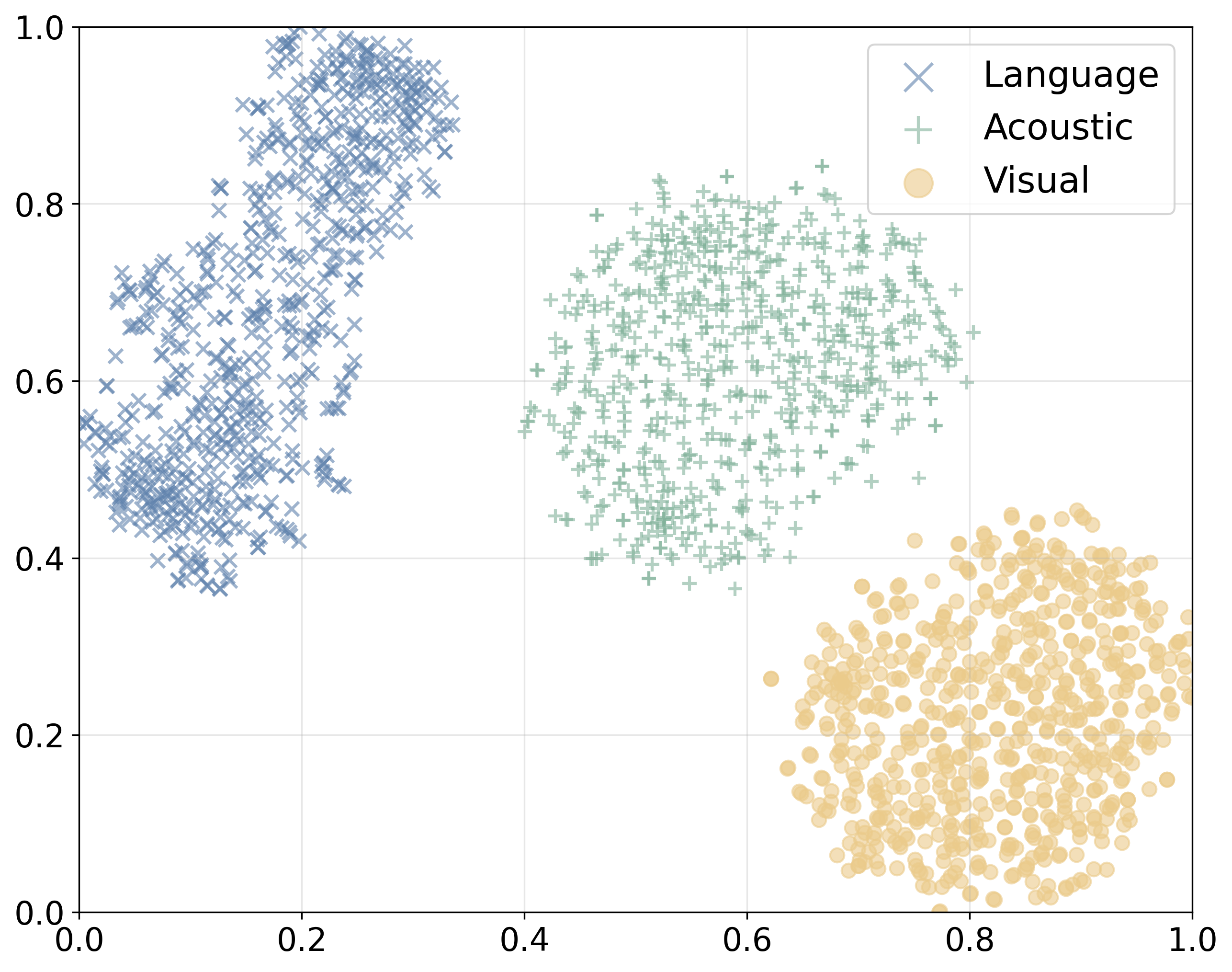}
        \caption{Diffusion Bridge }
    \end{subfigure}
    \begin{subfigure}[b]{0.19\linewidth}
        \centering
        \includegraphics[width=\linewidth]{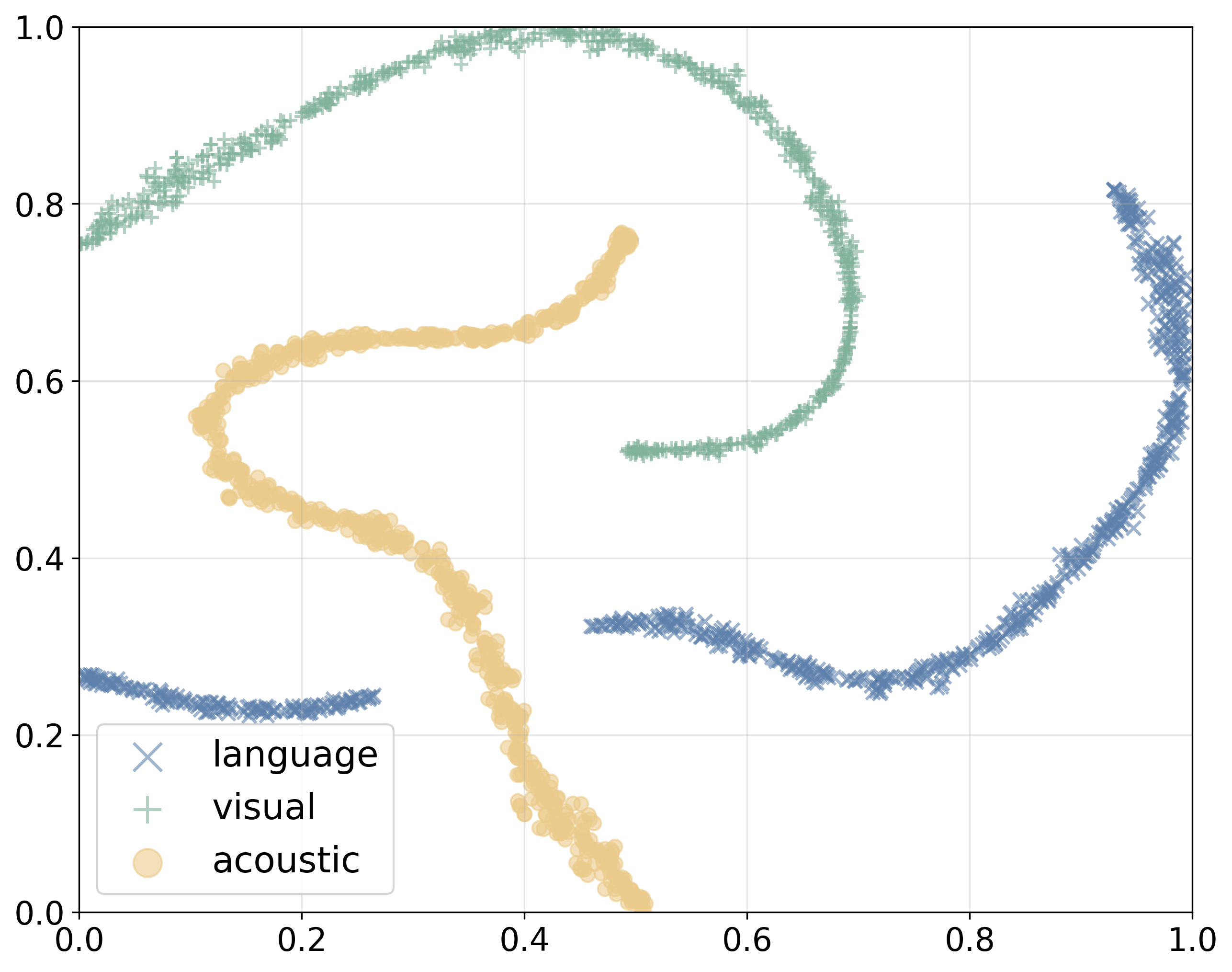}
        \caption{DLF (SOTA baseline) }
    \end{subfigure}
    \begin{subfigure}[b]{0.187\linewidth}
        \centering
        \includegraphics[width=\linewidth]{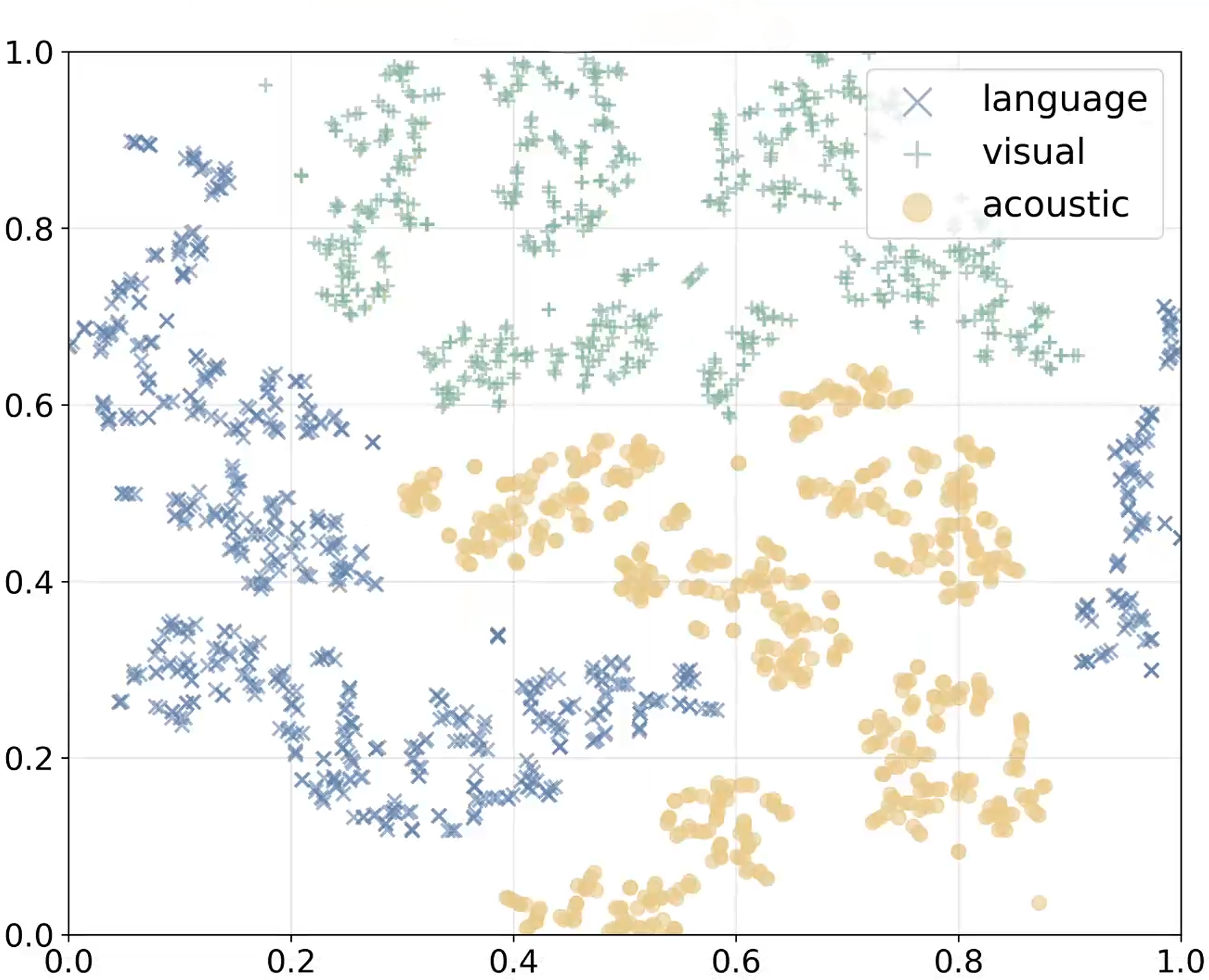}
        \caption{The proposed CaReFlow}
    \end{subfigure}
    \vspace{-0.2cm}
    \caption{The visualizations for distributions of unimodal features using t-SNE \cite{tsne}. CL denotes contrastive learning.}
    \label{9_vis}
    \vspace{-0.3cm}
\end{figure*}

\subsection{Discussion on Distribution Alignment Methods}

We compare CaReFlow with popular distribution mapping methods that are implemented under the same architecture as our CaReFlow for a fair comparison, including adversarial learning based model ARGF \cite{ARGF}, Deep CCA \cite{andrew2013deep}, contrastive learning based model CLGSI \cite{yang2024clgsi}, multimodal transformer (MulT) \cite{MULT}, and diffusion-based model Diffusion Bridge \cite{lee2025diffusion}. It can be inferred from Table~\ref{xxx} that CLGSI achieves the best performance among baselines. This is because CLGSI conducts one-to-many mapping between different modalities using contrastive learning, which can generate abundant modality pairs from different samples to learn a robust model that aligns modalities from the same class while pushing away modalities from different categories. Nevertheless, CaReFlow still outperforms CLGSI on all datasets. This is mainly because contrastive learning cannot perform adaptive alignment based on the degree of correlation between modalities, and cannot prevent information loss when aligning different modalities. In contrast, CaReFlow designs one-to-many mapping, adaptive relaxed alignment, and cyclic information flow to realize more accurate, robust, and information-preserving distribution alignment. In addition, in Section~\ref{sec:vis}, we conduct visualizations to demonstrate that CaReFlow can reduce modality gap more effectively compared to competitive baselines.

As for model complexity, as shown in Table~\ref{xxx},  the number of parameters of CaReFlow is slightly more than ARGF and Deep CCA, but less than CLGSI, MulT, and Diffusion Bridge.
This is because the only trainable modules introduced by CaReFlow are drift force models that are implemented as simple MLP networks, which have few parameters. The results indicate that \textbf{CaReFlow achieves the best performance with a moderate number of parameters}. The improvement in performance of CaReFlow does not come from the increase in the number of parameters, but from the enhancement of modality alignment effect.

\begin{figure*}
    \centering
    \begin{subfigure}[b]{0.24\linewidth} 
        \centering
        \includegraphics[width=\linewidth]{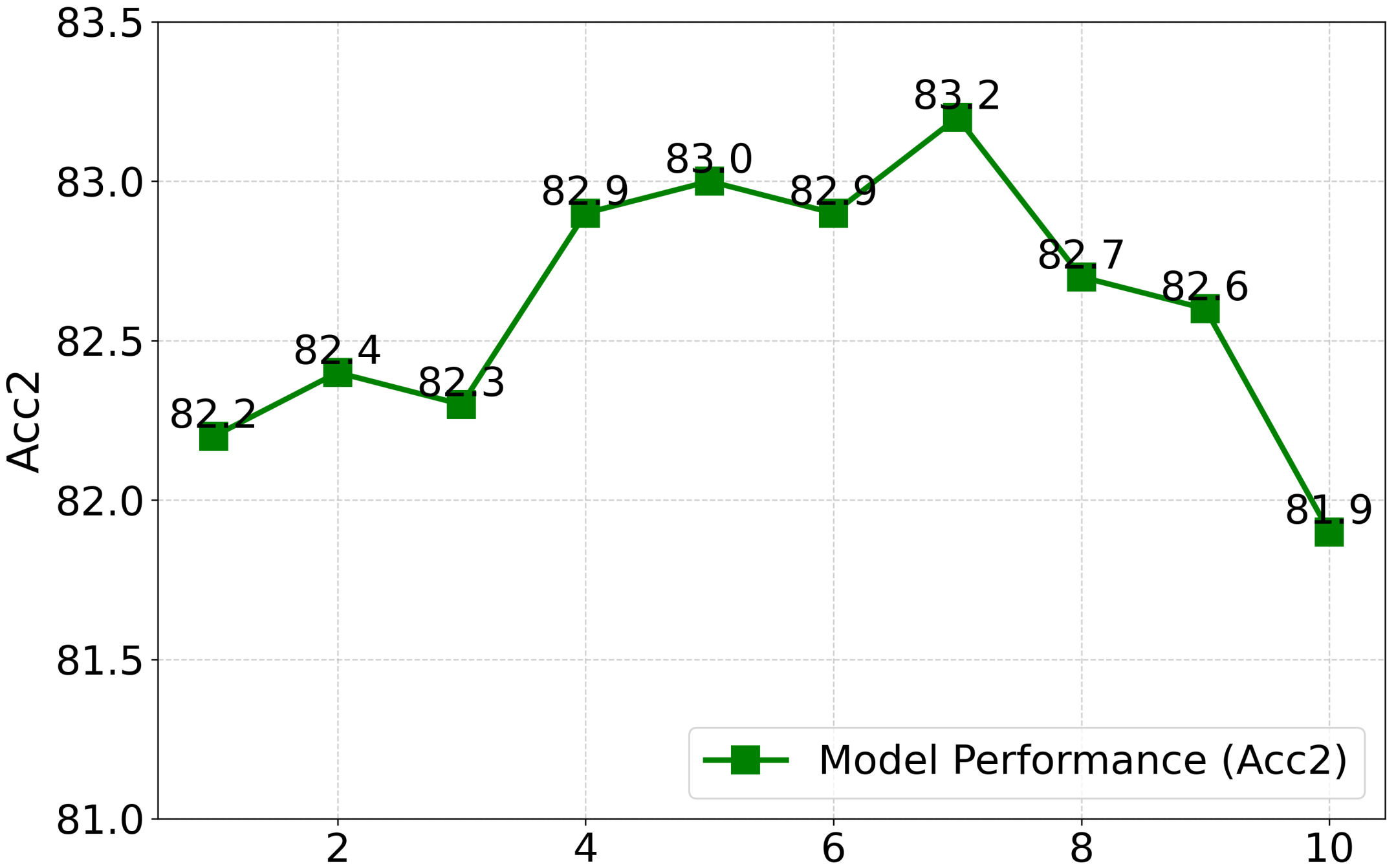}
        \caption{Sample Ratio $\beta$}
    \end{subfigure}
    \begin{subfigure}[b]{0.24\linewidth}
        \centering
        \includegraphics[width=\linewidth]{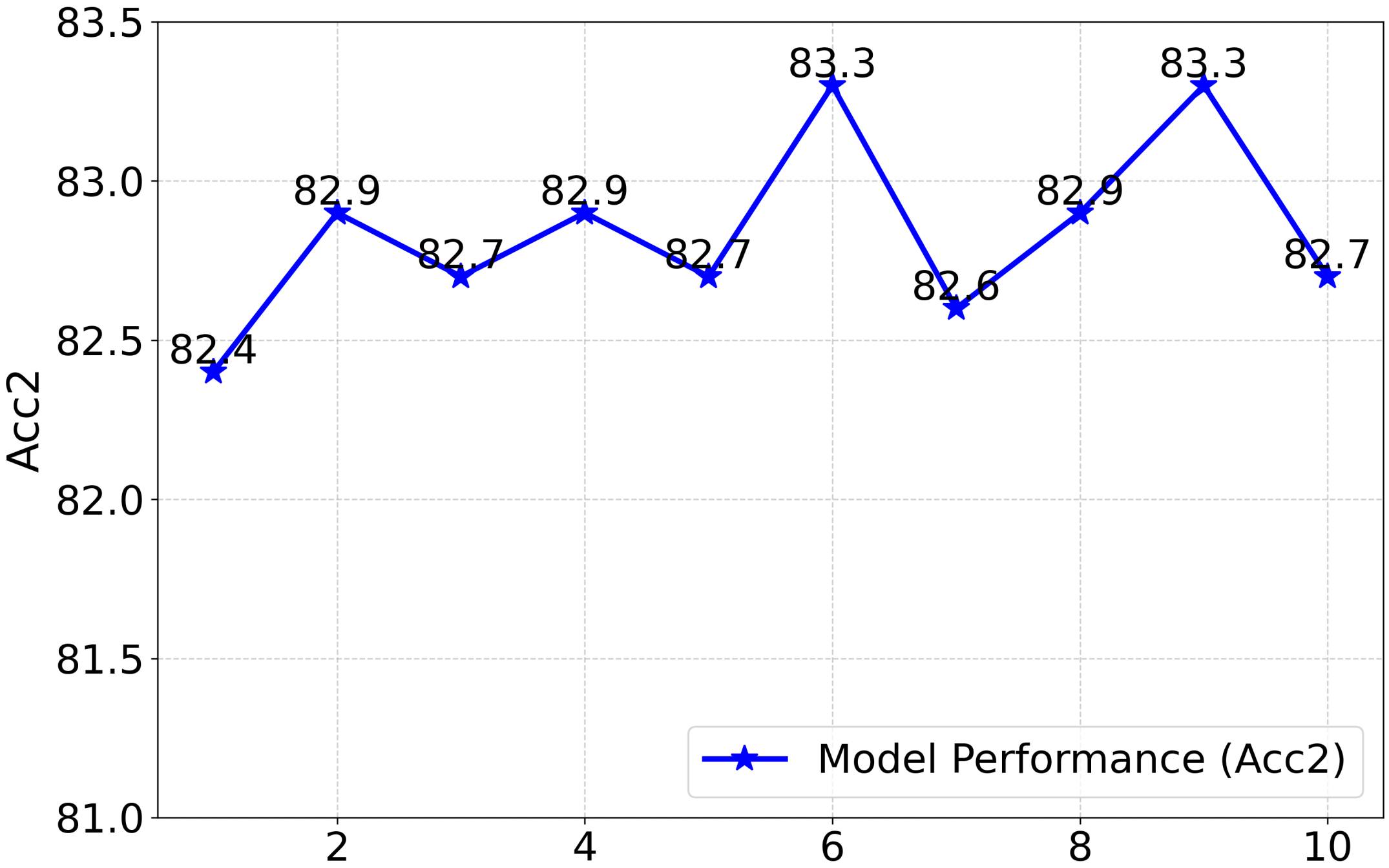}
        \caption{Number of Euler Steps $\frac{1}{dt}$}
    \end{subfigure}
    \begin{subfigure}[b]{0.24\linewidth}
        \centering
        \includegraphics[width=\linewidth]{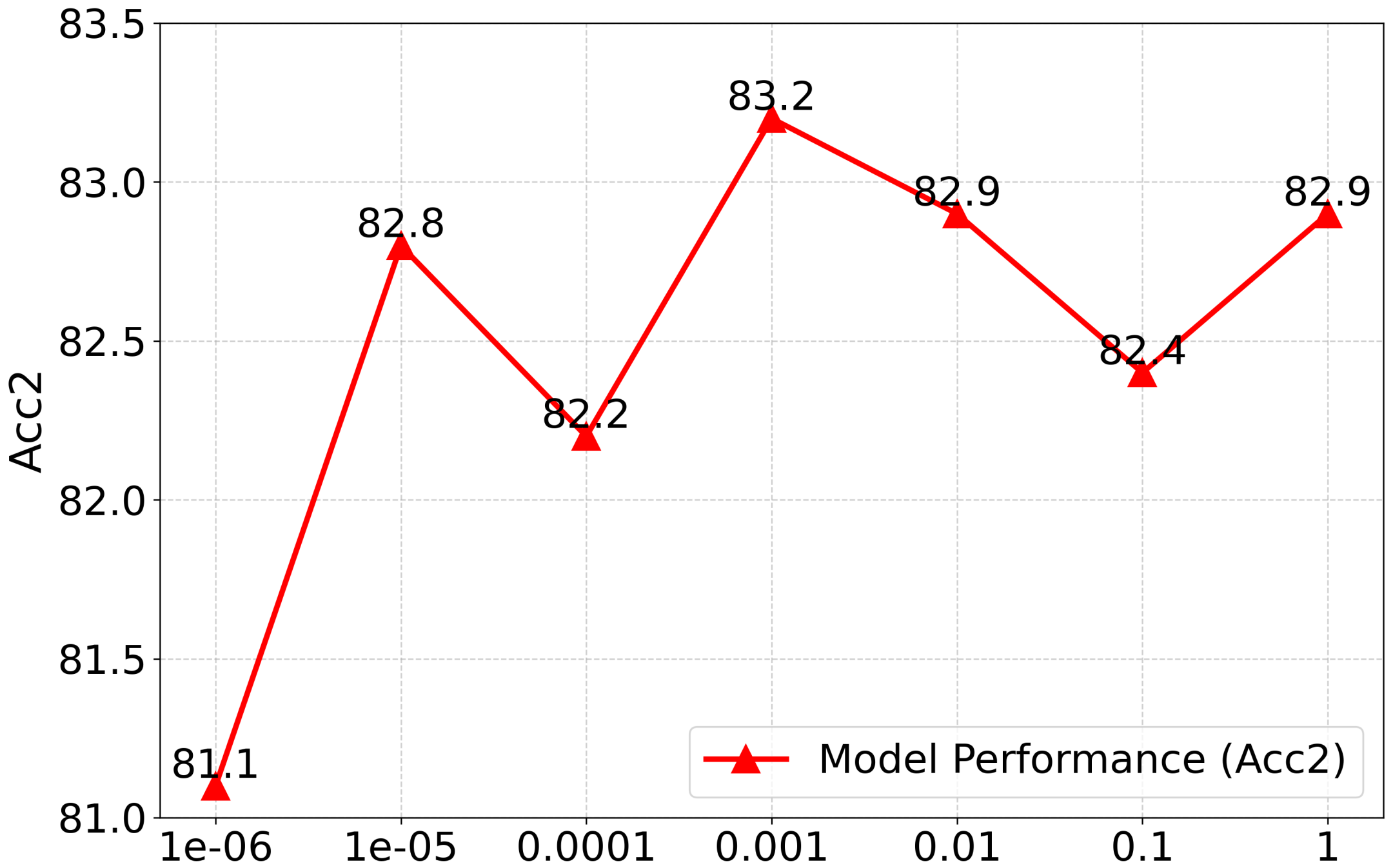}
        \caption{Forward Loss Weight $\alpha_f$}
    \end{subfigure}
    \begin{subfigure}[b]{0.24\linewidth}
        \centering
        \includegraphics[width=\linewidth]{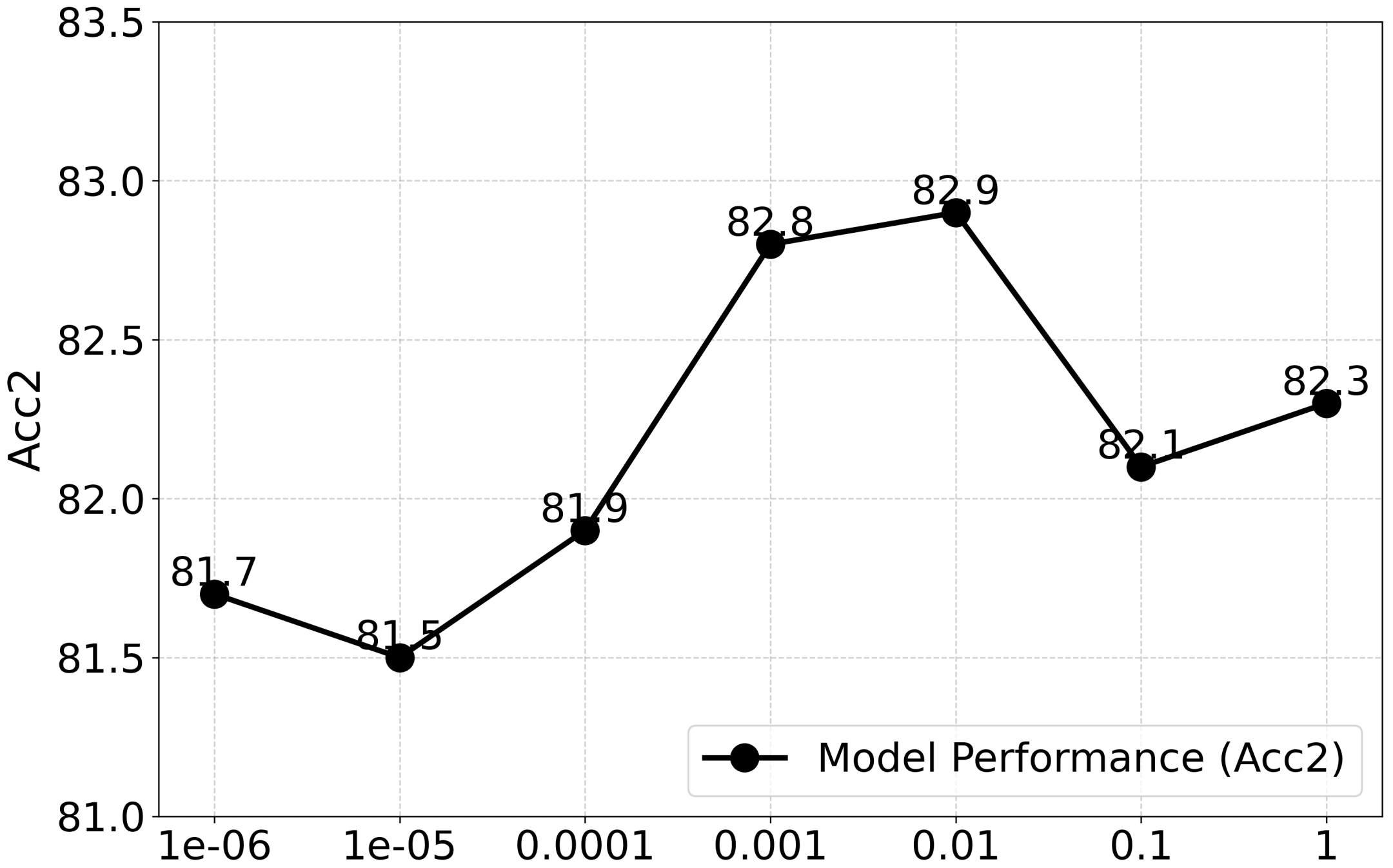}
        \caption{Backward Loss weight $\alpha_b$}
    \end{subfigure}
    \vspace{-0.2cm}
    \caption{The Acc2 w.r.t the change of sample ratio $\beta$, number of Euler steps $\frac{1}{dt}$, forward loss weight $\alpha_f$, and backward loss weight $\alpha_b$.}
    \label{9_hyper}
    \vspace{-0.3cm}
\end{figure*}

\subsection{Visualization on Unimodal Distributions} \label{sec:vis}

To verify that CaReFlow can effectively reduce the modality gap, we present the visualizations of unimodal features on CMU-MOSI. We also visualize the features of baselines for comparison, including three representative distribution mapping methods ARGF \cite{ARGF}, CLGSI \cite{yang2024clgsi} and Diffusion Bridge \cite{lee2025diffusion} as well as state-of-the-art baseline DLF \cite{wang2025dlf}. 
As shown in Figure~\ref{9_vis}, to our surprise, although the three distribution mapping methods have brought the features of different modalities closer in the feature space compared to the vanilla multimodal model (see Figure~\ref{vis_intro}), the reduction in the modality gap is not very significant. This may be because these methods typically train the main task and the distribution alignment task together, with the main task usually having a higher weight, potentially leading to insufficient optimization of the distribution alignment task. Compared to these three methods and DLF \cite{wang2025dlf}, \textbf{CaReFlow reduces the modality gap more effectively}, with a significant decrease in the distance between different modalities in the feature space. This is mainly because CaReFlow achieves one-to-many mapping and adaptive relaxed alignment, enabling robust and accurate learning of straight distribution transformation trajectory. Moreover, although the main task and distribution alignment task of CaReFlow are also trained simultaneously, CaReFlow uses the drift force model and detach operation to learn velocity vector, which to some extent decouples these two tasks and thus allows the distribution alignment task to be fully learned.

\subsection{Generalibility on Different Fusion Methods}

As demonstrated in Section~\ref{sec:pipeline}, CaReFlow uses a simple fusion network to achieve very competitive results. As CaReFlow is independent of fusion mechanisms, \textbf{we can incorporate advanced fusion mechanisms into CaReFlow to improve the performance}. To verify our statement, we investigate the performance of CaReFlow with various fusion methods, including tensor fusion, graph fusion, low-rank modality fusion, etc. We can infer from Table~\ref{t_fusion} that, compared to the original simple fusion network, more advanced fusion mechanisms achieve better results, indicating the potential of CaReFlow. Among all fusion methods, tensor fusion obtains the best results because of its high expressive power, which requires more parameters to process.

\begin{table}[t]
\centering
 \caption{ \label{t_fusion}Comparison of fusion methods on CH-SIMS-v2.
 }
 \vspace{-0.2cm}
\resizebox{.99\columnwidth}{!}{\begin{tabular}{c||c|c|c|c|c|c}
 \noalign{\hrule height 1pt} 
\rowcolor{lightgray!40}
    Fusion Methods & Acc5$\uparrow$ & Acc3$\uparrow$ & Acc2$\uparrow$ & F1$\uparrow$ & MAE$\downarrow$ & Corr$\uparrow$ \\
 \hline
 \hline
 Tensor Fusion \cite{Zadeh2017Tensor} & \textbf{58.6} & \textbf{77.9} &  \textbf{83.6} & \textbf{83.6}  & \underline{0.280} & \textbf{0.754} \\
 Graph Fusion \cite{ARGF} & \underline{58.1} & \underline{76.0} & 83.1  &  83.1  &  0.283 & \underline{0.749} \\
 Low-rank Modality Fusion \cite{Liu2018Efficient}   & 58.0 & 75.9 &  82.6 & 82.7 & 0.286 & 0.747 \\
   CubeMLP \cite{cubemlp_mm2022} & \underline{58.1} & 75.6 &  \underline{83.2} &  \underline{83.3} & \textbf{0.277} & 0.748 \\
 \hline
 \rowcolor[HTML]{EBFAFF}
Simple Fusion Network (Default) & 57.9 & 75.9 & 82.9 & 82.9 & \textbf{0.277} & 0.745  \\  
  \noalign{\hrule height 1pt}
 \end{tabular}}
 \vspace{-0.3cm}
\end{table}%

\subsection{Hyperparameter Robustness Analysis} \label{sec:hyper}

We evaluate the effect of critical hyperparameters on CH-SIMS-v2:
(a) \textbf{Sample Ratio} $\beta$: $\beta$ is a hyperparameter in one-to-many mapping strategy, denoting how many times the number of modality pairs from different samples is compared to those from the same sample. As shown in Figure~\ref{9_hyper} (a), $\beta$ should be a moderate value. This is mainly because when $\beta$ is too small, there are not enough training pairs, making it difficult to learn a robust drift force model. Conversely, when $\beta$ is too large, modality pairs from different samples may dominate the training, overshadowing the role of pairs from the same sample, and leading to difficulties in learning more accurate modality mapping relationships; (b) \textbf{Number of Euler Steps} $\frac{1}{dt}$: It represents the number of Euler steps required to transition from $\bm{X}_{m_1}$ to $\bm{X}_{m_2}$. As shown in Figure~\ref{9_hyper} (b), despite fluctuations in the number of Euler steps, the performance  varies very little, indicating its robustness. This implies that the transition trajectory learned by CaReFlow after single-shot training is relatively straight, because the performance of CaReFlow is satisfactory even when $\frac{1}{dt}$ is 2. This is mainly because we design adaptive relaxed alignment that allows CaReFlow to focus more on the target data points with greater correlation degrees when performing one-to-many mapping, thus learning more accurate modality distribution mapping in a faster way; (c) \textbf{Forward Loss Weight} $\alpha_f$ and \textbf{Backward Loss Weight} $\alpha_b$: $\alpha_f$ and  $\alpha_b$ denote the weights of forward and backward losses in CaReFlow, respectively. As shown in Figure~\ref{9_hyper} (c) and (d), $\alpha_f$ should be a large number. When $\alpha_f$ is small, forward loss does not exert its full effect, and CaReFlow fails to adequately transform source distributions to target distributions, leading to suboptimal fusion. The value of $\alpha_b$, however, should be a moderate number. 
This is because backward loss will dominate the training of the forward drift force model when $\alpha_b$ is large, causing it to focus solely on retaining the information of source modality. As a result, it cannot effectively transfer modality distribution.

Notably, \textbf{CaReFlow consistently outperforms baselines when hyperparameters are set to a variety of values, indicating its robustness} (see Table~\ref{tab:SIMSResult}). Moreover, when the hyperparameters are set to specific values (\eg, when $\beta$ is 7), the performance improves compared to when using default values (i.e., $\beta=4$), indicating \textbf{with more refined hyperparameter tuning, the results of CaReFlow can be further enhanced}, demonstrating its potential.

\section{Conclusion}

We propose CaReFlow to improve multimodal fusion via reducing the modality gap. CaReFlow leverages rectified flow to map the distribution of source modality to target modality via a `one-to-many' mapping strategy that strengthens the robustness of modality alignment.
In addition, CaReFlow designs adaptive relaxed alignment and cyclic information flow to achieve more accurate and information-preserving modality mapping. It achieves superior results on multiple MAC tasks with a simple fusion method and can effectively reduce the modality gap.

{
    \small
    \bibliographystyle{ieeenat_fullname}
    \bibliography{main}
}


\end{document}